%% file: neurips_2021_ml4ad.tex
\documentclass{article}

\PassOptionsToPackage{numbers}{natbib}

 \usepackage[final]{neurips_2021_ml4ad}

\usepackage[utf8]{inputenc} % allow utf-8 input
\usepackage[T1]{fontenc}    % use 8-bit T1 fonts
\usepackage{hyperref}       % hyperlinks
\usepackage{url}            % simple URL typesetting
\usepackage{booktabs}       % professional-quality tables
\usepackage{amsfonts}       % blackboard math symbols
\usepackage{nicefrac}       % compact symbols for 1/2, etc.
\usepackage{microtype}      % microtypography
\usepackage{xcolor}         % colors

\usepackage{graphicx}
\usepackage{subcaption}
\usepackage[frozencache=true,cachedir=./anc]{minted}
\usepackage{placeins}
\usepackage{paralist}
\usepackage{nameref}
\usepackage{amsmath}
\usepackage{amssymb}

\usepackage{bbold}

\newcommand{\SKIP}[1]{}

\newcommand{\bfx}{\mathbf{x}}

\newcommand{\bfy}{\mathbf{y}}

\newcommand{\bfs}{\mathbf{s}}
\newcommand{\bfz}{\mathbf{z}}

\newcommand{\dataset}{\mathcal{D}}

\newcommand{\calX}{\mathcal{X}}
\newcommand{\calY}{\mathcal{Y}}
\newcommand{\calZ}{\mathcal{Z}}
\newcommand{\calS}{\mathcal{S}}

\newcommand{\indicator}{\mathbb{1}}
\newcommand{\expectation}{\mathbb{E}}

\title{A Step Towards Efficient Evaluation of Complex Perception Tasks in Simulation}

\author{%
  Jonathan Sadeghi \\
  Five AI Ltd. \\
  \texttt{jonathan.sadeghi@five.ai}
  \And
  Blaine Rogers \\ 
  Five AI Ltd. \\
  \And
  James Gunn \\
  Five AI Ltd. \\
  \And
  Thomas Saunders \\
  Five AI Ltd. \\
  \And
  Sina Samangooei \\ 
  Five AI Ltd. \\
  \And
  Puneet Kumar Dokania \\
  Five AI Ltd. \\
  \And
  John Redford \\
  Five AI Ltd. \\
}

\begin{document}

\maketitle

\begin{abstract}

There has been increasing interest in characterising the error behaviour of systems which contain deep learning models before deploying them into any safety-critical scenario.
However, characterising such behaviour usually requires large-scale testing of the model that can be extremely computationally expensive for complex real-world tasks.
For example, tasks involving compute intensive object detectors as one of their components.
In this work, we propose an approach that enables efficient large-scale testing using simplified low-fidelity simulators and without the computational cost of executing expensive deep learning models.
Our approach relies on designing an efficient surrogate model corresponding to the compute intensive components of the task under test. 
We demonstrate the efficacy of our methodology by evaluating the performance of an autonomous driving task in the Carla~\cite{Dosovitskiy17} simulator with reduced computational expense by training efficient surrogate models for PIXOR~\cite{yang2018pixor} and CenterPoint~\cite{yin2020center} LiDAR detectors, whilst demonstrating that the accuracy of the simulation is maintained.

\end{abstract}

\input{introduction}

\input{methodology}

\input{related_work}

\input{experiments}

\input{conclusions}

\FloatBarrier
\begin{ack}
We are grateful to all colleagues at Five who have contributed to insightful discussions about Perception Error Models, including Philip Torr, Andrew Blake, Simon Walker, John McDermid, Sebastian Kaltwang, Adam Charytoniuk, Torran Elson and Romain Mueller. 
We are particularly grateful to Romain Mueller, Luca Bertinetto, Francesco Pinto and Anuj Sharma for proofreading this manuscript.
\end{ack}

{\small
\bibliographystyle{plainnat} 
\bibliography{bibliography}
}

%\clearpage
\appendix
\input{appendix}
\end{document}

%% file: introduction.tex
\section{Introduction}
Deep learning models achieve state of the art performance for a variety of real-world applications~\cite{janai2020computer}, however the fact that these models have multiple vulnerabilities such as shift in data distribution~\cite{ovadia2019can} and additive perturbations~\cite{eykholt2018robust, ilyas2019adversarial, madry2018towards, tsipras2019robustness}, has limited their practical usability in safety-critical situations such as driverless cars. 
A solution to this problem is to collect and annotate a large diverse dataset that captures all possible scenarios for training and testing. 
However, since the costs involved in manually annotating such a large quantity of data can be prohibitive, it is often beneficial to employ high-fidelity simulators~\cite{johnson2017driving} to potentially produce vast number of diverse scenarios with exact ground-truth annotations at almost no cost.

\begin{figure}[h]
\centering
\includegraphics[width=0.99\linewidth]{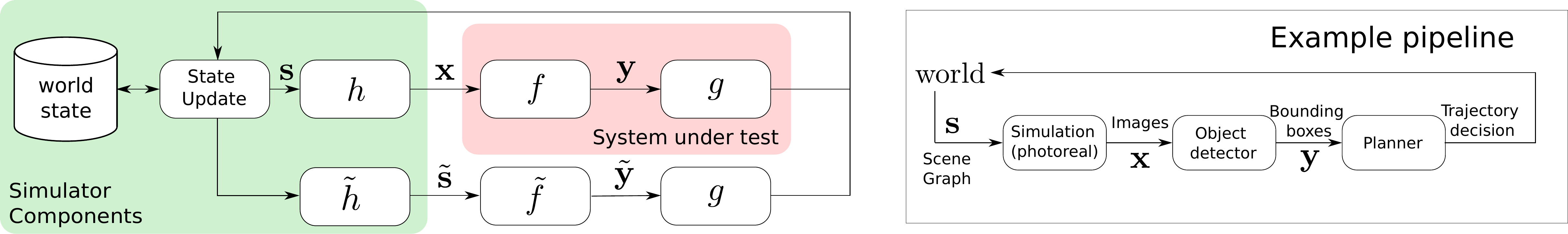}
\caption{
Diagram to show how the simulator, $h$, and backbone task $f$ can be circumvented during end-to-end testing with a surrogate model $\tilde{f}$.
The outputs of the surrogate model and the backbone task are fed into the model for the downstream task $g$ in the same way because $\bfy$ and $\tilde{\bfy}$ are the elements of the same vector space.
The simulator will update the world state $\bfs$ based on the output of the downstream task $g$, which is used by the expensive simulator $h$ to produce sensor readings $\bfx$, and efficient low-fidelity simulator $\tilde{h}$ to produce the low-dimensional inputs to the surrogate model $\tilde{\bfs}$. 
}
\label{fig:prism}
\end{figure}

We would like to use the abundance of annotated samples provided by a simulator to perform \emph{extensive testing} of a given deep learning model.
This will allow us to find failure modes of such models before deployment.
For example, let us assume that our objective is to find failure modes of a path planner (downstream task $g$) of a driverless car that takes detected objects as an input from a trained object detector (backbone task $f$)~\cite{board2019collision, janai2020computer}. 
Since the failure modes of the detector would have significant impact on the planner, we would like to test the planner by giving inputs directly
to the detector~\cite{tampuu2020survey, afzal2020study}, and thereby capture dependencies between the detector and the planner. 
In this setting, one might want to use all possible high-fidelity synthetic generations $\bfx$ from the simulator as an input to the detector to characterise the failure modes of the planner.
These failure modes can be identified using metrics for the planner behaviour to highlight data sequences which should be inspected by the engineer.
However, given that the inference step of an object detector where the input is
high-fidelity synthetic data itself is a computationally demanding operation \cite{tan2020efficientdet}, such an approach will not be scalable enough to perform extensive testing of the task.

Motivated by the above example, the goal in this work is to propose an efficient alternative to testing with a high-fidelity simulator and thereby enable large-scale testing.
Our approach is to replace the computationally demanding backbone task $f$ with an efficient surrogate $\tilde{f}$ that is trained to mimic the \textit{behaviour} of the backbone task. 
As opposed to $f$ where the input is a high-fidelity sample ${\bf x}$, the input to the surrogate is a much lower-dimensional `salient' variable $\tilde{\bfs}$. 
In the example of the object detector and path planner: $\bfx$ might be a high-fidelity simulation of the output of camera or LiDAR sensors, whereas $\tilde{\bfs}$ might be simply the position, orientation and level of occlusion of other vehicles and pedestrians in the scene together with other aspects of the scene like the level of lighting which could also affect the results of the detection function $f$. 
The training of the surrogate is performed to provide the following approximate composite model $g(\tilde{f}(\tilde{\bfs})) \approx g(f(\bfx))$. 
This allows us to efficiently perform rigorous testing on the downstream task using very low-dimensional inputs to an efficient surrogate model for the backbone, as shown in Figure~\ref{fig:prism}. 
This results in a reduction in compute time whilst maintaining the benefits of using a simulator for testing.

In summary, our contributions are as follows:
\begin{compactitem}
	\item We propose a surrogate modelling approach to enable efficient large-scale testing of complex tasks, and describe one specific methodology for creating such surrogate models in the autonomous driving setting.
    \item We perform extensive large-scale experiments demonstrating the efficacy of our approach using Carla~\cite{Dosovitskiy17} simulator for driving tasks. 
    We create surrogate models of two well-known LiDAR detectors, PIXOR~\cite{yang2018pixor} and Centerpoint~\cite{yin2020center}, as our backbone task.
    \item We show that our approach is closest to the backbone task compared to the baselines evaluated on several metrics, and yields a 20 times reduction in the compute time.
\end{compactitem}

%% file: methodology.tex
\section{Methodology}
\label{sec:methodology}

Training and testing a model for a complex task primarily involves the following three modules--(1) a {\bf data generator} $(h)$ that provides the high-fidelity sensory input domain $\calX$; (2) a {\bf backbone-task} $(f_{\theta})$, parameterised by $\theta$, that maps $\bfx \in \calX$ into an intermediate representation $\bfy \in \calY$; and (3) a {\bf downstream task} $(g_{\phi})$, parameterised by $\phi$, that takes the intermediate $\bfy$ as an input and maps it into a desired output $\bfz \in \calZ$. For example, devising a path planner that takes as input the raw sensory data $\bfx$ from the world via sampler $h$ and outputs an optimal trajectory would rely on intermediate solutions such as accurate detections $f_{\theta}(\bfx)$ by an object detector $f$ (backbone task) in order to provide the optimal trajectory $g_{\phi}(f_{\theta}(\bfx))$. Most real-world problems consist of such complex tasks that heavily depend on intermediate solutions, obtaining which can sometimes be the main bottleneck from both efficiency and accuracy points of view. Considering the same example of a path planner, it is well-known that an object detector is computationally expensive in real-time. Therefore, extensively evaluating the planner that depends on the detector can quickly become computationally infeasible as there exist millions of diverse road scenarios over which the planner should be tested before deployment into any safety-critical environment such as driverless cars in the real-world.

We identify the two following bottlenecks of evaluating extensively such complex tasks: (1) efficiently obtaining a sufficiently diverse set of test scenarios; and (2) efficient inference of the intermediate expensive backbone tasks. 
Though simulators, as used in this work, can theoretically solve the first problem as they can provide infinitely many test scenarios, their use, in practice, is limited as it still is very expensive for the backbone task $f$ to process these high-fidelity samples obtained from the simulator, and for the simulator to generate these samples.

Our solution to alleviate these bottlenecks is rather simple. 
Instead of obtaining high-fidelity samples from the simulator $h$, we generate samples in a low-dimensional vector space which is designed so that these embeddings summarise the crucial information required by the backbone task $f$ to be able to provide accurate predictions. 
Using these low-dimensional simulator outputs, a rather simplistic and efficient model $\tilde{f}$ can then be trained to mimic the behaviour of the target backbone task $f$ which provides the input for the complex composite task $g$ under test.
This will allow very fast and efficient evaluation of the composite task by approximating the inference of the backbone task using a surrogate model.
In what follows, we provide details of these approximations.

\paragraph{Obtaining Low-fidelity Simulation Data:} The data generation process for simulators is a mapping $h:\bfs \mapsto \bfx$, where $\bfs \in \calS$ denotes the world-state that is normally structured as a scene graph (refer Figure~\ref{fig:prism}). Note, for high-fidelity generations, $\bfs$ is very high dimensional as it contains all the properties of the world necessary to generate a realistic sensor reading $\bfx$. For example, in the case of road scenarios, it contains, but is not limited to, positions and types of all vehicles, pedestrians and other moving objects in the world, details of the road shapes and surfaces, surrounding buildings, light and RADAR reflectivity of all surfaces in the simulation, and lighting and weather conditions~\cite{Dosovitskiy17}. There is usually a trade-off between an accurate simulator and the computational expense of the simulator. Therefore, even if a high fidelity simulator is available, it may be intractable to produce sufficient simulated data for training and evaluation as the mapping $h$ itself is expensive.

Noting that a low-fidelity simulator $\tilde{h} : \bfs \mapsto \tilde{\bfs}$ can be created to map high-dimensional $\bfs$ into low-dimensional `salient' variables $\tilde{\bfs} \in \tilde{\calS}$ for a variety of tasks~\cite{pouyanfar2019roads}, in this work, we use $\tilde{\bfs}$ as an input to the surrogate backbone task (design of the surrogate is discussed below).
In the simplest case, the mapping $\tilde{h}(.)$ could consist of a subsetting operation. For example, in object detectors, $\tilde{h}$ could output $\tilde{\bfs}$ that contains the position and size of all the actors in the scene. In order to provide more useful information in $\tilde{\bfs}$, we also include \textit{low-fidelity physical simulations} in $\tilde{h}$. Specifically, we use a ray tracing algorithm in $\tilde{h}$ to calculate the geometric properties such as occlusion of actors in the scene for one of the ego vehicle's sensors. 
Clearly, the subsetting operation and deciding what physical simulations to include in $\tilde{h}$ requires domain knowledge about the backbone task.
We believe this is not an unreasonable assumption as for most of the perception related tasks that we are interested in, we have a fairly good idea of what factors are necessary to capture the underlying performance. A more generic setting to automatically learn $\tilde{h}$ could be an interesting future direction.

\paragraph{Efficient Surrogate for the Backbone Task:}\label{para:surrogate_design} The next step is to use the low-dimensional $\tilde{\bfs}$ in order to provide reliable inputs for the downstream task. Recall, our objective is to provide an {\em efficient way to mimic} $f_{\theta}(h(\bfs))$ so that its output can be passed to the downstream task for large-scale testing (refer Figure~\ref{fig:prism}). To this end, using the low-dimensional $\tilde{\bfs}$ as input, we design and train a surrogate function $\tilde{f}_{\tilde{\theta}}$ such that $\tilde{f}_{\tilde{\theta}} (\tilde{\bfs}) \approx f_{\theta}(h(\bfs))$, for all $\bfs \in \calS$. By design, the surrogate function takes a very low-dimensional input compared to the high-fidelity $\bfx$ and, as will be shown in our experiments, is orders of magnitude faster than operating a high-fidelity simulator, $h$, and the original backbone task, $f$.
The time required for training the surrogate model is typically insignificant compared to the overall simulation budget, and therefore is not discussed in further detail.
The performance of the downstream task may differ when the backbone task is replaced with the surrogate model, due to the approximate nature of the surrogate model; erroneous outputs of the surrogate model could be predicted by a full treatment of model uncertainty for the surrogate model, however we do not explore this issue further in the present work.

We now provide the details of the surrogate model. As mentioned, the selection of the salient variables $\tilde{\bfs}$ and the form of the surrogate function is a design choice that requires domain knowledge. 
Since our experimentation in this work primarily involves testing of a planner that requires an object detector as the backbone task, our choice of salient variables for the input to the detector surrogate includes: {\em position, linear velocity, angular velocity, actor category, actor size, and occlusion percentage} (we will specify in each experiment exactly which variables were used). 
Additional salient variables could also be used, but as a proof of concept of our approach, we chose the above salient variables and did not explore more combinations as these variables already provided promising results. 
To compute the occlusion percentage, we simulate a low-resolution semantic LiDAR and calculate the proportion of rays terminating in the desired agent's bounding box~\cite{xiang2013object}. 
Typically, these salient variables are available at no computational cost when the simulator updates the world-state.

We design a simple probabilistic neural network $\tilde{f}_{\tilde{\theta}}$ 
as the surrogate for the object detector. 
To train $\tilde{f}$, for every $\bfs$, a tuple $\{ \tilde{\bfs} = \tilde{h}(\bfs), \bfy = f(h(\bfs)) \}$ of input-output is created for every frame, which we process to obtain an input-output tuple for each agent in the scene. 
Running the perception system to create the training data for the surrogate model incurs an additional computational cost, however we do not consider this to be problematic because the cost is fixed for a particular perception system, and can therefore be amortised over simulations using many different planners.
We use the Hungarian algorithm with an intersection over union cost between objects \cite{forsyth2012computer} to associate the ground-truth locations and the detections from the {\em original} backbone task $f$, on a per-frame basis. 
Therefore, the training data for the surrogate detector would be $\dataset = \{\tilde{\bfs}^i, \tilde{\bfy}^i\}_{i=1}^k$, and although we have defined $\tilde{f}$ as a function of all ground-truth objects in the scene, in this paper our implementation factorises over each agent in the scene and acts on a single agent basis.
This results in a surrogate model which by design cannot predict False Positive detections, as is the case in most of the perception error model literature (see \cite{piazzoni2020modeling, piazzonimodeling, hirsenkorn2016virtual, zec2018statistical, krajewskineural}).
Choosing a more general surrogate model would avoid this limitation, however, we consider this out of the scope of the present work.
Our network architecture for the surrogate is a multi-layered fully-connected network with skip connections, and dropout layers between `skip blocks' (similar to a ResNet \cite{he2016deep}), which is shown in the Appendix~\ref{sec:architecture}.
The final layer of the network outputs the parameters of the underlying probability distributions, which are usually a Gaussian distribution (mean and log standard deviation) for the detected position of the objects, and a Bernoulli distribution for the binary valued outputs, e.g. objectness for prediction of false negatives~\cite{krajewskineural}.
The training is performed by maximizing the following expected log-likelihood:
%\vspace{-0.5em}
\begin{equation}
    \mathcal{L}_\text{total} = \sum_i \log p(\tilde{\bfy}_\text{det}^{i}|\tilde{\bfs}^{i}) + \indicator_{\{\bfy_\text{det}^{i} = 1\}} \log p(\tilde{\bfy}_\text{pos}^{i}| \tilde{\bfs}^{i}),
    \label{eqn:loss}
\end{equation}
%\vspace{-0.5em}
where, associated with the surrogate function $\tilde{f}_{
\tilde{\theta}}(.)$, $p(\cdot|\tilde{\bfs}^{i})$ represents the likelihood, $\tilde{\bfy}_\text{det}$ represents the Boolean output which is true if the object was detected, and $\tilde{\bfy}_\text{pos}$ represents a real-valued output describing the centre position of the detected object, respectively.
The term $\log p(\tilde{\bfy}_\text{det}^{(i)}|\tilde{\bfs}^{(i)})$ in Eqn.~\ref{eqn:loss} is equivalent to the binary cross-entropy when using a Bernoulli distribution to predict false negatives. Assuming Cartesian components of the positional error to be independent, we obtain:
%\vspace{-0.5em}
\begin{equation}
\begin{aligned}
    \log p(\tilde{\bfy}_\text{pos}^{i}| \tilde{\bfs}^{i})  
    = \log{\left(\mathcal{N}(\tilde{\bfy}_\text{x, pos}^{(i)}; \mu_x, \sigma_x) \cdot \mathcal{N}(\tilde{\bfy}_\text{y, pos}^{(i)}; \mu_y, \sigma_y) \right)} \\
    = \frac{(\tilde{\bfy}_\text{x, pos}^{(i)} - \mu_x)^2}{2\sigma_x^2} + \frac{(\tilde{\bfy}_\text{y pos}^{(i)} - \mu_y)^2}{2\sigma_y^2} + \log \sigma_x \sigma_y
\end{aligned}
\label{eqn:gaussian_loss}
\end{equation}
%\vspace{-0.5em}
where $\mu$ and $\log{(\sigma)}$ are the outputs of the fully connected neural network.
For further details on how probabilistic neural networks can be trained for multiple target variables, see \cite{kendall2017uncertainties, kendall2018multi}.

%% file: related_work.tex
\section{Related work}
\label{sec:related_work}

\paragraph{End-to-end evaluation in simulation:} 
End-to-end evaluation refers to the concept of evaluating components of a modular machine learning pipeline together in order to understand the performance of the system as a whole.
Such approaches often focus on strategies to obtain equivalent performance using a lower fidelity simulator whilst maintaining accuracy to make the simulation more scalable \cite{pouyanfar2019roads, balakrishnan2020closing, el2019LiDAR}.
Similarly, \citet{urtasun} use a realistic LiDAR simulator to modify real-world LiDAR data which can then be used to search for adversarial traffic scenarios to test end-to-end autonomous driving systems.
\citet{kadian2019we} attempt to validate a simulation environment by showing that an end-to-end point navigation network behaves similarly in the simulation to in the real-world by using the correlation coefficient of several metrics in the real-world and the simulation.
End-to-end testing is possible without a simulator, for example \citet{philion2020learning} evaluate the difference between planned vehicle trajectories when planning using ground truth and a perception system and show that this enables important failure modes of the perception system to be identified.
End to end testing could be performed more efficiently using knowledge distillation or network quantisation \cite{cheng2017survey}. 
However, such approaches would still require running $h$ to produce $\bfx$, which is typically computationally expensive and therefore we do not perform further comparison with these approaches.

Our approach differs from these in that the surrogate model methodology enables end-to-end evaluation without running the backbone model in the simulation.

\paragraph{Perception error models (PEMs):} 
Perception Error Models (PEMs) are surrogate models used in simulation to replicate the outputs of perception systems so that downstream tasks can be evaluated as realistically as possible.
\citet{piazzoni2020modeling} present a PEM for the pose and class of dynamic objects, where the error distribution is conditioned on the weather variables, and use the model to validate an autonomous vehicle system in simulation on urban driving tasks.
\citet{piazzonimodeling} describe a similar approach using a time dependent model and a model for false negative detections.
Time dependent perception PEMs have also been used by \citet{berkhahntraffic} to model traffic intersections with a behaviour model and a stochastic process misperception model on velocity, and \citet{hirsenkorn2016virtual} by creating a Kernel Density Estimator model of a filtered radar sensor, where the simulated sensor is modelled by a Markov process.
\citet{zec2018statistical} propose to model an off the shelf perception system using a Hidden Markov Model.
Modern machine learning techniques have also been used to create PEMs, for example \citet{krajewskineural} create a probabilistic neural network model for a LiDAR sensor, \citet{arnelid2019recurrent} use Recurrent Conditional Generative Adversarial Networks to model the output of a fused camera and radar sensor system, and \citet{suhre2018simulating} describe an approach for simulating a radar sensor using conditional variational auto-encoders.

Perception Error Models should not be confused with surrogate models which approximate the entire system simulation, i.e. $g(f(h(\bfs))$ over all time steps, rather than the perception system specifically.
These full-system surrogate models are used in simulation to efficiently search for failures in autonomous driving systems \cite{Beglerovic2017, sinha2020neural, corso, uesato2018rigorous}.

Our model is most similar to \citet{mitra2018towards}, where an autoregressive neural network PEM is evaluated in simulation by visually comparing ego behaviour to the behaviour under no perception errors.
We describe a more general framework for the training of surrogate models in a probabilistic machine learning context with a larger-scale evaluation than the other papers in this section.

%% file: experiments.tex
\section{Experiments}
\label{sec:experiment}
\paragraph{Overview:}
We use Carla simulator \cite{Dosovitskiy17} to analyze the behaviour of an agent in two driving tasks: (\textbf{1}) adaptive cruise control (ACC) and (\textbf{2}) the Carla leaderboard.
The Carla configuration is provided in Appendix~\ref{sec:sim_config}.
The agent uses a LiDAR object detector $f$ (backbone task) to detect other agents and make plans accordingly.
Using the methodology described in Section~\ref{sec:methodology}, we construct a \textbf{Neural Surrogate (NS)} $\tilde f$ that, as opposed to $f$, does not depend on high-fidelity simulated LiDAR scans.
We show that the surrogate agents behave similarly to the real agent while being extremely efficient. 

\paragraph{Details of Driving Scenarios:}
\begin{compactenum}
    \item \emph{Adaptive Cruise Control (ACC)}: The agent follows a fast moving vehicle. Suddenly, the fast moving vehicle cuts out into an adjacent lane, revealing a parked car. The agent must brake to avoid a collision. See \cite{jurgen2006adaptive} for a review of ACC.
    \item \emph{Carla Leaderboard}: See \citet{carlaChallenge} for details. This contains a far more diverse set of driving scenarios, including urban and highway driving, and is approximately a two orders of magnitude increase in the total driving time relative to the ACC task. Therefore, this evaluation can be seen as a large scale evaluation of our methodology.
\end{compactenum}
For the ACC task we use a very simple planner described in Section~\ref{sec:sim_experiment} that maintains a constant speed and brakes to avoid obstacles. For the more demanding Carla Leaderboard we use a more robust planner and detector, described in detail in Section~\ref{sec:leaderboard_evaluation}.

\paragraph{Baselines:}
We compare our approach Neural Surrogate (NS) against three strong baseline surrogate models ($\tilde{f}$), which are similar to those used in the PEM literature described in Section~\ref{sec:related_work}:
\begin{compactitem}
    \item Planning using \emph{ground truth (GT)} locations of agents, which are available from $\tilde{\bfs}$.
    \item A \emph{logistic regression (LR)} surrogate is trained to predict the false negatives (missed detections) of the backbone task $f$ for a given $\tilde{\bfs}$. Note, $\tilde{\bfs}$ here is same as the salient variables used in NS. The logits for the true class probability are then predicted by first passing the input via a linear mapping $W$ and then applying a sigmoid function, i.e. $p_t = \sigma(W \tilde{\bfs})$ where $\sigma$ is the sigmoid function. The model is trained using the focal loss, $\mathcal{L}(p_t)=-\alpha_t (1-p_t)^\gamma \log (p_t)$ where $p_t$ represents the true class probability and $\alpha_t={\indicator_{\{\bfy_\text{det}^{i} = 1\}}} \alpha + \indicator_{\{\bfy_\text{det}^{i} = 0\}}(1-\alpha)$, in order to account for class imbalance \cite{puneet_focal_loss, lin2017focal}. The hyperparameters used for the focal loss were manually tuned to be $\alpha=0.6$ and $\gamma=2$ via cross-validation over the classification metrics.
    Once a box has been identified as a true-positive detection of $f$ by the LR model, its \textbf{exact} co-ordinates from $\tilde{\bfs}$ are then passed to the planner.
    \item A \emph{Gaussian Fuzzer (GF)} is a simple surrogate model where the exact position and velocity of a box obtained from $\tilde{\bfs}$ are simply \textbf{perturbed} with samples from independent Gaussian and StudentT distributions respectively (StudentT distributions are used due to the heavy tails of the detected velocity errors). This is Eqn.~\ref{eqn:gaussian_loss} with fixed $\mu$ and $\sigma$, i.e. not a function of other variables in $\tilde{\bfs}$. These parameters are obtained analytically using Maximum Likelihood Estimation (MLE). For example, for the Gaussian distribution over positional errors, the MLE solution is simply the empirical mean and the standard deviation of the detection position errors which is obtained using the train set \cite{bishop2013pattern}.
\end{compactitem}
In the Carla leaderboard evaluation, only the ground truth baseline is used.
The hyperparameters used for the training of all the surrogate models are shown in Appendix~\ref{sec:hyperparams}.

\paragraph{Surrogate Training Data:}
In both experiments, the Carla leaderboard scenarios are used to obtain training data for surrogate models and the LiDAR detector.
The dataset consists of driving scenarios which are approximately 10 minutes long, featuring a variety of vehicles, cyclists and pedestrians in a mixture of urban and highway scenes.
\begin{compactenum}
    \item In the first experiment, scenarios 0-9 are used for training and scenario 10 is used for testing. Pedestrians are excluded from the data because the ACC task does not involve pedestrians.
    \item The second experiment contains a wider variety of driving scenarios so the collected dataset is larger; scenarios 0-58 are used for training and scenarios 59-62 are used for testing.
\end{compactenum}

\paragraph{Metrics:}
We use common classification and regression metrics to directly compare the outputs of the surrogate model and the real model on the backbone task:
\begin{compactitem}
    \item Precision and recall.
    \item Classification accuracy.
    \item Sampled position mean squared error (spMSE): the mean squared error of the detections $\bfy$ or surrogate predictions $\tilde{\bfy}$, as appropriate, relative to the ground truth values in $\bfs$.
\end{compactitem}

We wish to quantify (1) how closely the surrogate mimics the backbone task $f$; and (2) how close it is to the ground-truth obtained from $\tilde{\bfs}$.
While evaluating a surrogate model \emph{relative to $f$}, a false negative of $\tilde{f}$ would be a situation when an agent is detected by $\tilde{f}$ which was in fact missed; conversely evaluating a surrogate model \emph{relative to the ground truth} means that a false negative of $\tilde{f}$ would be when an agent is not detected by $\tilde{f}$ which is in fact present in the ground truth data.  
When we evaluate surrogate models \emph{relative to the detector} (comparing $\bfy$ to $\tilde{\bfy}$), the best surrogate is the one with the highest value of the evaluation metric. 
However, when we evaluate surrogate models \emph{relative to the ground truth} (comparing $\bfy$ or $\tilde{\bfy}$, as appropriate, to $\bfs$), the best surrogate is the one whose score is closest to the detector's score. 
The metrics are only evaluated for objects within 50m of the ego vehicle, since objects further than this are unlikely to influence the ego vehicle's behaviour.

We use other metrics to compare the performance of the surrogate and real agents on the downstream task.
For the ACC task we evaluate the runtime per frame (with and without $h$ or $\tilde{h}$), Maximum Braking Amplitude (MBA), and MBA timestamp (tMBA) relative to the start of the scenario. MBA quantifies the degree to which the braking was applied relative to the maximum possible braking.

We also evaluate the mean Euclidean norm (meanEucl), which we define as the time integrated norm of the stated quantity, i.e. to compare variables $v_1(t)$ and $v_2(t)$, the metric is
\begin{equation}
	\frac{1}{t'} \int_{t=0}^{t=t'} ||v_1(t) - v_2(t)||_2 dt. \label{eqn:euclidean}
\end{equation}
We evaluate Eqn.~\ref{eqn:euclidean} with a discretised sum.
This metric is a natural, time dependent, method of \textit{comparing trajectories} in Euclidean space.
In Appendix~\ref{sec:pkl-comparison}, we provide a relationship between Eqn.~\ref{eqn:euclidean}, and the planner KL-divergence metric proposed by \citet{philion2020learning}.
We also compute the maximum Euclidean norm (maxEucl) to show the maximum instantaneous difference in the stated quantity, 
\begin{equation}
	\max_{t\in[0,t']} ||v_1(t) - v_2(t)||_2 . \label{eqn:euclidean_max}
\end{equation}
In the Carla leaderboard task we compare the metrics used for Carla leaderboard evaluation i.e. \emph{route completion, pedestrian collisions and vehicle collisions} for the detector and NS.
We also compute the cumulative distribution functions of the time between collisions for the detector, NS, and GT.

\subsection{Carla Adaptive Cruise Control Task}
\label{sec:sim_experiment}

\paragraph{Configuration Details:}
In this experiment, our backbone task $f$ consists of a PIXOR LiDAR detector \cite{yang2018pixor} trained on simulated LiDAR pointclouds from Carla \cite{yang2018pixor}, followed by a Kalman filter which enables the calculation of the velocity of objects detected by the LiDAR detector.
Therefore, $\bfy$ and $\tilde{\bfy}$ consist of position, velocity, agent size and a binary variable representing detection of the object.
To simplify the surrogate model, in this particular experiment, we assume that the ground-truth value of the agent size can be used by the planner whenever required.
The salient variables $\tilde{\bf{s}}$ consist of position, orientation, velocity, angular velocity, object extent, and percentage occlusion.
The downstream task consists of a planner which is shown in further detail in Appendix~\ref{sec:planner_code}.

\paragraph{Results:}
Regression and classification performance \textit{relative to the ground-truth} on the train and test set are shown in Table~\ref{tab:test} for both the surrogate models and the detector.
Table~\ref{tab:test_comparitive} shows similar metrics to Table~\ref{tab:test}, but this time computed for the surrogate models \textit{relative to the detector}.
This shows that although the LR surrogate is predicting a similar proportion of missed detections, the NS is more effective at predicting these when the detector would also have missed the detection.
Appendix~\ref{sec:error_distributions} contains plotted empirical cumulative distribution functions for the positional and velocity error predicted by each surrogate model and the true detector relative to the true object locations.

\begin{table}
\caption{Metrics comparing surrogates  $\tilde{f}$ (LR, NS, and GF), and the PIXOR detector (the backbone task $f$) on the train and test set \textit{relative to the ground-truth}. The surrogate closest to the detector is shown in bold. Note, surrogates trivially achieve a precision of 1 as they do not model false positives.}
\label{tab:test}
\centering
\scalebox{0.82}{
\begin{tabular}{@{}l@{\hspace{8pt}}l@{\hspace{4pt}}l@{\hspace{4pt}}l@{\hspace{4pt}}l@{\hspace{4pt}}l@{}}
			\toprule
			 & NS (our) & GF & LR & Detector \\
			\midrule
			
			& \multicolumn{4}{c}{Train Set} \\
			\cmidrule{2-5}
			Precision & \textbf{1} & \textbf{1} & \textbf{1} & 0.934 \\

			Recall & \textbf{0.642} & 1 & 0.218 & 0.614 \\
		
			spMSE & \textbf{0.231} & 0.294 & 0 & 0.239 \\
			\bottomrule
			\end{tabular}
			\begin{tabular}{@{}l@{\hspace{4pt}}l@{\hspace{4pt}}l@{\hspace{4pt}}l@{\hspace{4pt}}l@{}}
			\toprule
			 NS (our) & GF & LR & Detector \\
			\midrule
			 \multicolumn{4}{c}{Test Set} \\
			\cmidrule{1-4}
			\textbf{1} & \textbf{1} & \textbf{1} & 0.842 \\
		
			 \textbf{0.533} & 1 & 0.238 & 0.476 \\
	
			 \textbf{0.267} & 0.295 & 0 & 0.273 \\
		\bottomrule
		\end{tabular}
		}
\end{table}

\begin{table}
\centering
\caption{Metrics comparing LR, NS (our) and GF surrogate models on the train and test set \textit{relative to the detector}. The GF trivially achieves a recall of 1, since the detector does not model false negatives.}
\scalebox{0.82}{
\begin{tabular}{@{}l@{\hspace{8pt}}l@{\hspace{4pt}}l@{\hspace{4pt}}l@{\hspace{4pt}}l@{}}
            \toprule
			 & LR & GF & NS (our)  \\
			\midrule
			&\multicolumn{3}{c}{Train Set} \\
			\cmidrule{2-4}
			Accuracy $\uparrow$ & 0.688 & 0.244 & \textbf{0.954} \\

			Recall $\uparrow$ & 0.277 & \textbf{1} & 0.931 \\
			
			Precision $\uparrow$ & 0.332 & 0.244 & \textbf{0.886} \\
            \bottomrule
            \end{tabular}
\begin{tabular}{@{}l@{\hspace{4pt}}l@{\hspace{4pt}}l@{\hspace{4pt}}l@{}}          
            \toprule
             LR & GF & NS (our)  \\
			\midrule
			\multicolumn{3}{c}{Test Set} \\
			\cmidrule{1-3}
			 0.784 & 0.176 & \textbf{0.915} \\
		
			  0.412 & \textbf{1} & 0.823 \\
			
			 0.392 & 0.176 & \textbf{0.730} \\
			\bottomrule
\end{tabular}
}
\label{tab:test_comparitive}
\end{table}

MBA and time efficiency results are shown in Table~\ref{tab:metrics_single}. The surrogates are multiple times faster than the backbone task while showing MBA behaviour similar to the backbone task. 
Notably the wall-time taken per step (DTPF) is about \textbf{100 times higher} for the PIXOR Detector than the surrogate models, excluding the simulator rendering time, with all models running on an Intel Core i7-8750H CPU.
Including the simulator rendering time, the difference is reduced to \textbf{20 times} (TTPF), indicating that the majority of the time savings are achieved by removing the object detector from the simulation pipeline.
The TTPF for GF is approximately 0.06 seconds less than for the other surrogate models since in this case the headless simulator $\tilde{h}$ does not have to calculate the occlusion of agents.
In real-world applications, detectors typically run on GPU hardware which results in much quicker inference. 
However, for simulation purposes, removing the requirement of using a GPU for perception inference has the potential to significantly reduce costs.

In Table~\ref{tab:metrics_pairwise}, a selection of pairwise metrics is shown comparing the ego trajectory in each simulation environment.
The pairwise metrics show that using a surrogate model produces closer agent behaviour to the backbone task (LiDAR detector) compared to GT, both for metrics based on velocity and position.
The NS is the best performing model on all pairwise metrics. Plots of the actors' trajectories are shown in Appendix~\ref{sec:diagnostic_plots}.
The GF produces similar ego trajectories to the GT baseline, and this is most likely because false negatives, which cause delayed braking and are therefore influential in this scenario, are not included in both cases.
The metrics indicate that the LR model is most similar to the NS, however, the ego trajectories produced by the LR are less similar to those produced by the LiDAR detector than those produced by the NS.

\begin{table}
\centering
\caption{
Timestamp of Maximum Braking Amplitude (tMBA), MBA, Detector time / frame for $f$ and $\tilde{f}$ (excluding simulator rendering time) (DTPF), and Total Time / Frame (TTPF) -- including simulator time ($h$, $\tilde{h}$), state update, and the planner time ($g$). For all metrics the most similar model to the detector is shown in bold. All time noted in seconds.
}
\scalebox{0.82}{
\begin{tabular}{@{}l@{\hspace{4pt}}l@{\hspace{4pt}}l@{\hspace{4pt}}l@{\hspace{4pt}}l@{\hspace{4pt}}l@{\hspace{4pt}}l@{}}
\toprule
  & Det. & GF & NS (our) & LR & GT \\
\midrule
			tMBA & 13.5 & 19.0 & \textbf{13.5} & 15.8 & 19.9 \\
	
			MBA & 1 & 0.378 & \textbf{1} & 0.518 & 0.093 \\
		
			DTPF & 1.511 & 0.013 & 0.017 & 0.013 & 0.005 \\
		
			TTPF & 1.644 & 0.023 & 0.087 & 0.083 & 0.022 \\
 \bottomrule
\end{tabular}
}
\label{tab:metrics_single}
\end{table}

\begin{table}
\caption{Mean and Max Euclidean norm of displacement, computed for the ego position and velocity traces separately (Eqn.~\ref{eqn:euclidean} and Eqn.~\ref{eqn:euclidean_max}). For both metrics the most similar model to the detector $f$ is shown in bold. The normalised value of the metric, obtained by dividing the metric by the (GT, Det). value, is shown in brackets.}
\centering

\scalebox{0.82}{
\begin{tabular}{@{}l@{\hspace{8pt}}l@{\hspace{4pt}}l@{\hspace{4pt}}l@{\hspace{4pt}}l@{\hspace{4pt}}l@{\hspace{4pt}}c@{}}
\toprule
   & Det. & GF & NS (our) & LR & GT \\
   \cmidrule{2-6}
   \multicolumn{6}{c}{meanEucl Position $\downarrow$} \\
 \cmidrule{2-6}
			Det. & 0 (0.0)& 4.9 (0.98)& \textbf{0.40 (0.08)} & 1.6 (0.32) & 4.9 (1.0) \\
		
			GF &  & 0 (0.0) & 4.5 (0.92) & 4.1 (0.83) & 0.15 (0.03) \\
			
			NS &  &  & 0 (0.0) & 1.2 (0.24) & 4.6 (0.93) \\
		
			LR &  &  &  & 0 (0.0) & 4.2 (0.85) \\
			
			GT &  &  &  &  & 0 (0.0) \\
 \cmidrule{2-6}
   \multicolumn{6}{c}{meanEucl Velocity $\downarrow$ } \\
 \cmidrule{2-6}
			Det. & 0 (0.0) & 1.5 (0.99) & \textbf{0.20 (0.13)} & 0.71 (0.46) & 1.5 (1.0) \\
			
			GF &  & 0 (0.0) & 1.3 (0.87) & 0.93 (0.60) & 0.027 (0.017) \\
		
			NS &  &  & 0 (0.0) & 0.52 (0.34) & 1.3 (0.88) \\
	
			LR &  &  &  & 0 (0.0) & 0.94 (0.61) \\
			
			GT &  &  &  &  & 0 (0.0) \\
 \bottomrule
 \end{tabular}
 \begin{tabular}{@{}l@{\hspace{4pt}}l@{\hspace{4pt}}l@{\hspace{4pt}}l@{\hspace{4pt}}l@{\hspace{4pt}}c@{}}
\toprule
   Det. & GF & NS (our) & LR & GT \\
   \midrule
   \multicolumn{5}{c}{maxEucl Position $\downarrow$ } \\
 \cmidrule{1-5}
			 0.0 (0.0) & 19 (0.99) & \textbf{2.2 (0.11)} & 7.9 (0.41) & 19 (1.0) \\
		
			  & 0.0 (0.0) & 17. (0.87) & 11. (0.57) & 0.36 (0.019) \\
		
			  &  & 0.0 (0.0) & 5.7 (0.30) & 17. (0.89) \\
		
			  &  &  & 0.0 (0.0) & 11. (0.59) \\
	
			  &  &  &  & 0.0 (0.0) \\

 \midrule
   \multicolumn{5}{c}{maxEucl Velocity $\downarrow$ }\\
 \cmidrule{1-5}
			 0.0 (0.0) & 5.5 (0.99) & \textbf{2.0 (0.35)} & 4.1 (0.73) & 5.5 (1.0) \\
	
			 & 0.0 (0.0) & 4.7 (0.84) & 3.4 (0.61) & 0.11 (0.019) \\
			
			 &  & 0.0 (0.0) & 3.2 (0.57) & 4.7 (0.85) \\
		
			 &  &  & 0.0 (0.0) & 3.4 (0.62) \\
		
			 &  &  &  & 0.0 (0.0) \\
\bottomrule
\end{tabular}
}
\label{tab:metrics_pairwise}
\end{table}

\subsection{Carla Leaderboard Evaluation}
\label{sec:leaderboard_evaluation}

\paragraph{Configuration Details:}
In this experiment, the backbone task $f$ is a CenterPoint LiDAR detector \cite{yin2020center} for both vehicles and pedestrians, trained on the simulated data from Carla in addition to proprietary real-world data. 
The downstream planner $g$ is a modified version of the \texttt{BasicAgent} from the Carla Python API, which follows a fixed path and performs emergency braking to avoid vehicle and pedestrian collisions (described in further detail in Appendix~\ref{sec:basicagent}).

The NS model architecture is mostly the same as in Section~\ref{sec:methodology}, but the agent velocity is removed from $\bfy$, since the \texttt{BasicAgent} does not require the velocities of other agents.
In addition, an extra salient variable is provided to the network in $\tilde{\bfs}$: a one hot encoding of the class of the ground truth object (vehicle or pedestrian) and in the case of the object being a vehicle, the make and model of the vehicle.
Since the training dataset is imbalanced and contains more vehicles at large distances from the ego vehicle, minibatches for the training are created by using a stratified sampling strategy: the datapoints are weighted using the inverse frequency in a histogram over distance with 10 bins, resulting in a balanced distribution of vehicles over distances.

\begin{figure}
     \centering
          \begin{subfigure}[b]{0.5\linewidth}
         \centering
         \includegraphics[width=\linewidth]{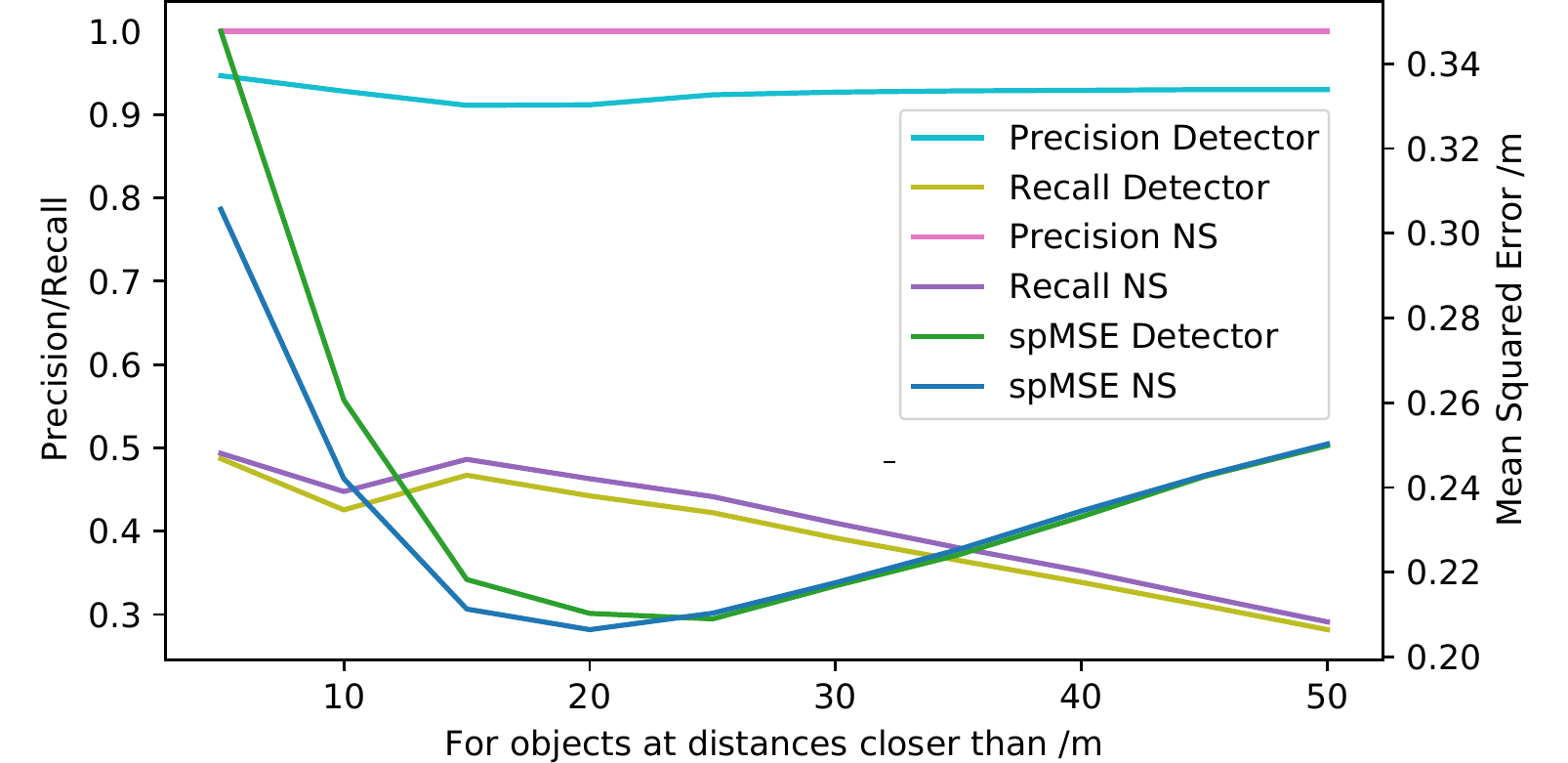}
         \caption{Train set metrics}
     \end{subfigure}\hfill
     \begin{subfigure}[b]{0.475\linewidth}
         \centering
         \includegraphics[width=\linewidth]{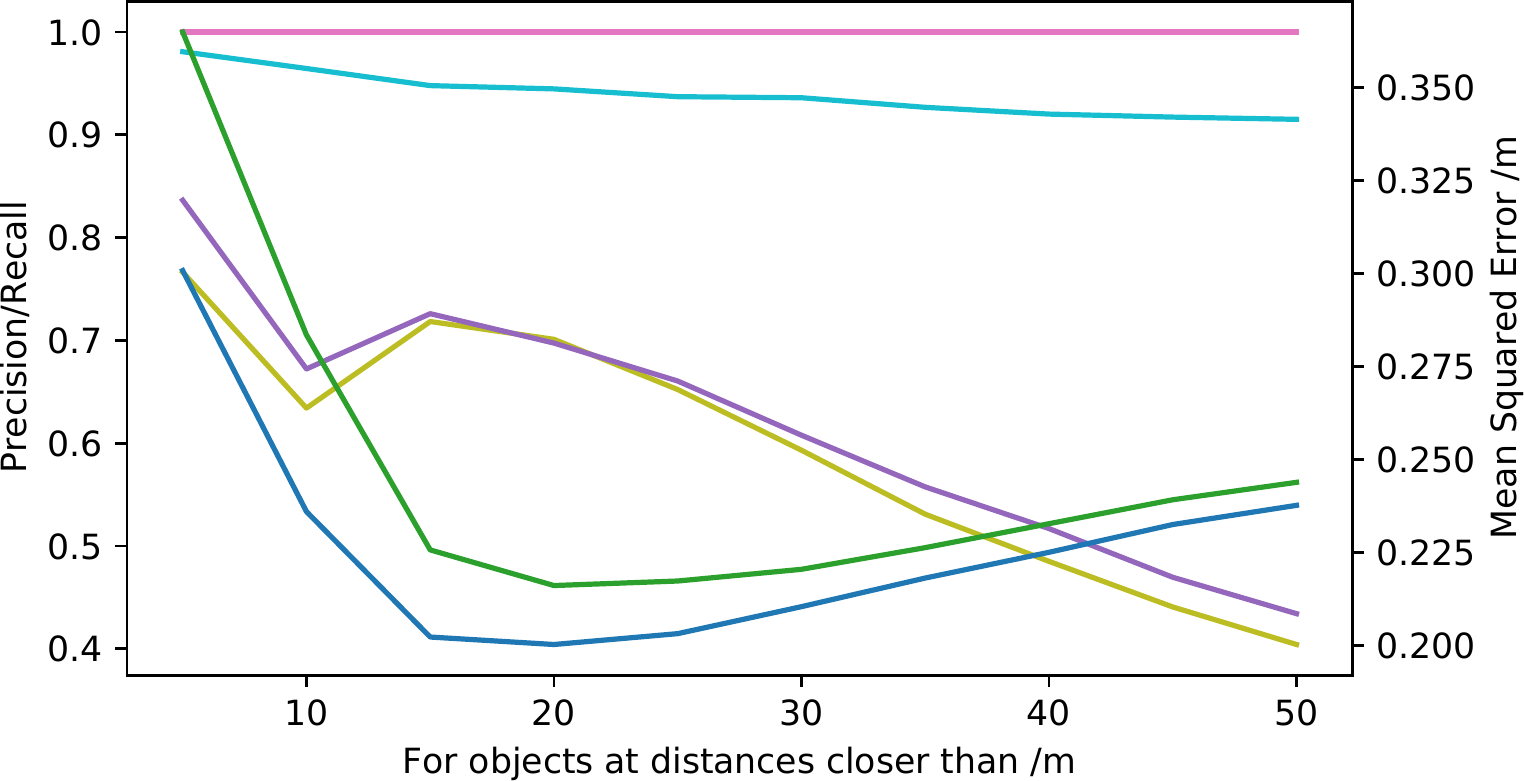}
         \caption{Test set metrics}
     \end{subfigure}
        \caption{Metrics comparing NS and the CenterPoint detector on train and test set \textit{relative to the ground-truth}. 
        The precision of NS is always 1 because, by design, it is unable to predict false positives (we do not re-associate the detections to ground truth after they are generated by the NS, and instead perform the evaluation using the association used to generate the boxes).
        }
	\label{fig:test_2}
\end{figure}

\begin{figure}
\begin{minipage}[t]{0.45\linewidth}
         \includegraphics[width=\linewidth]{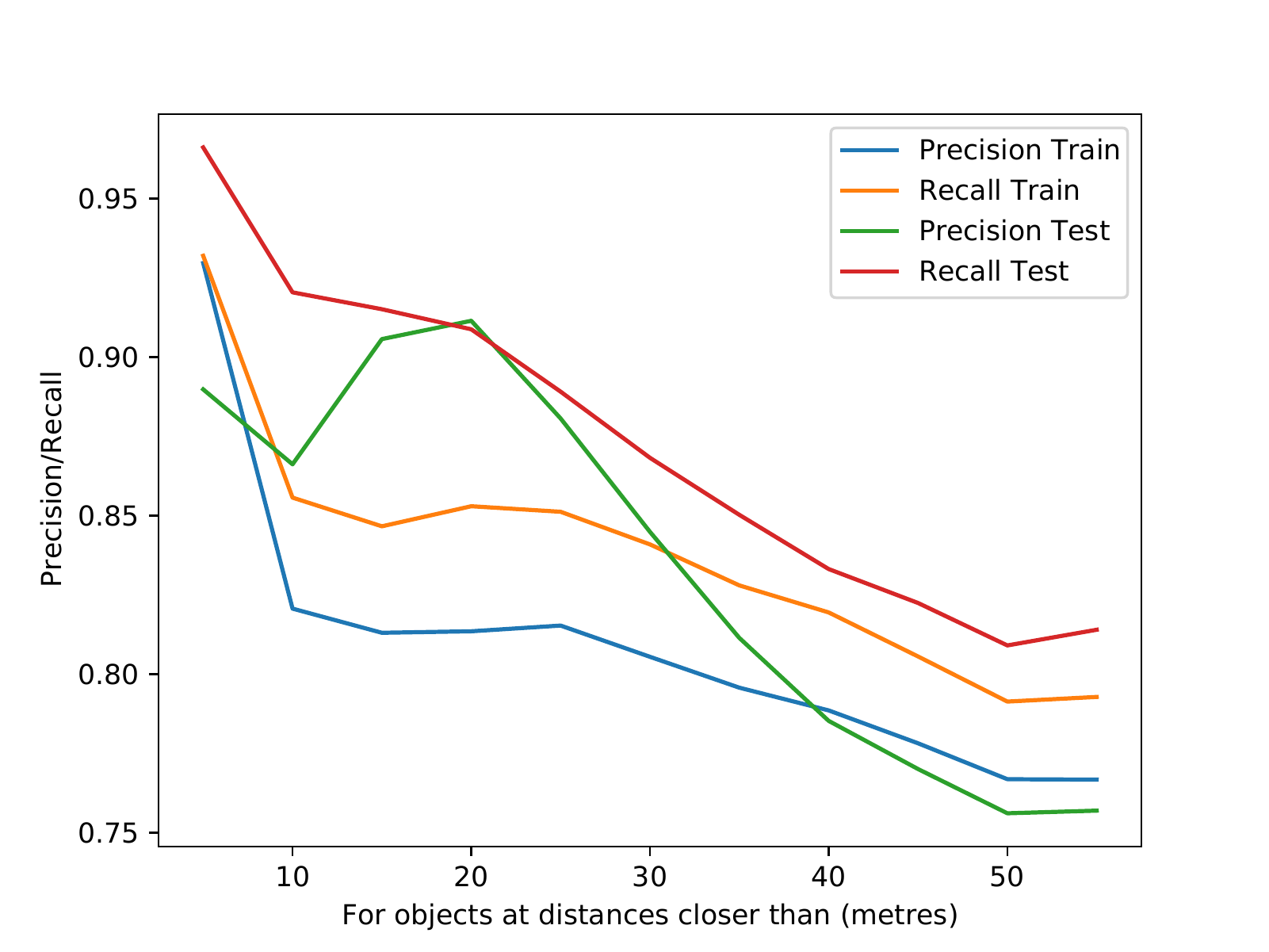}
\caption{Comparing NS \textit{relative to the detector}.}
\label{fig:test_comparitive_2}
\end{minipage}\hfill
\begin{minipage}[t]{0.45\linewidth}
         \includegraphics[width=\linewidth]{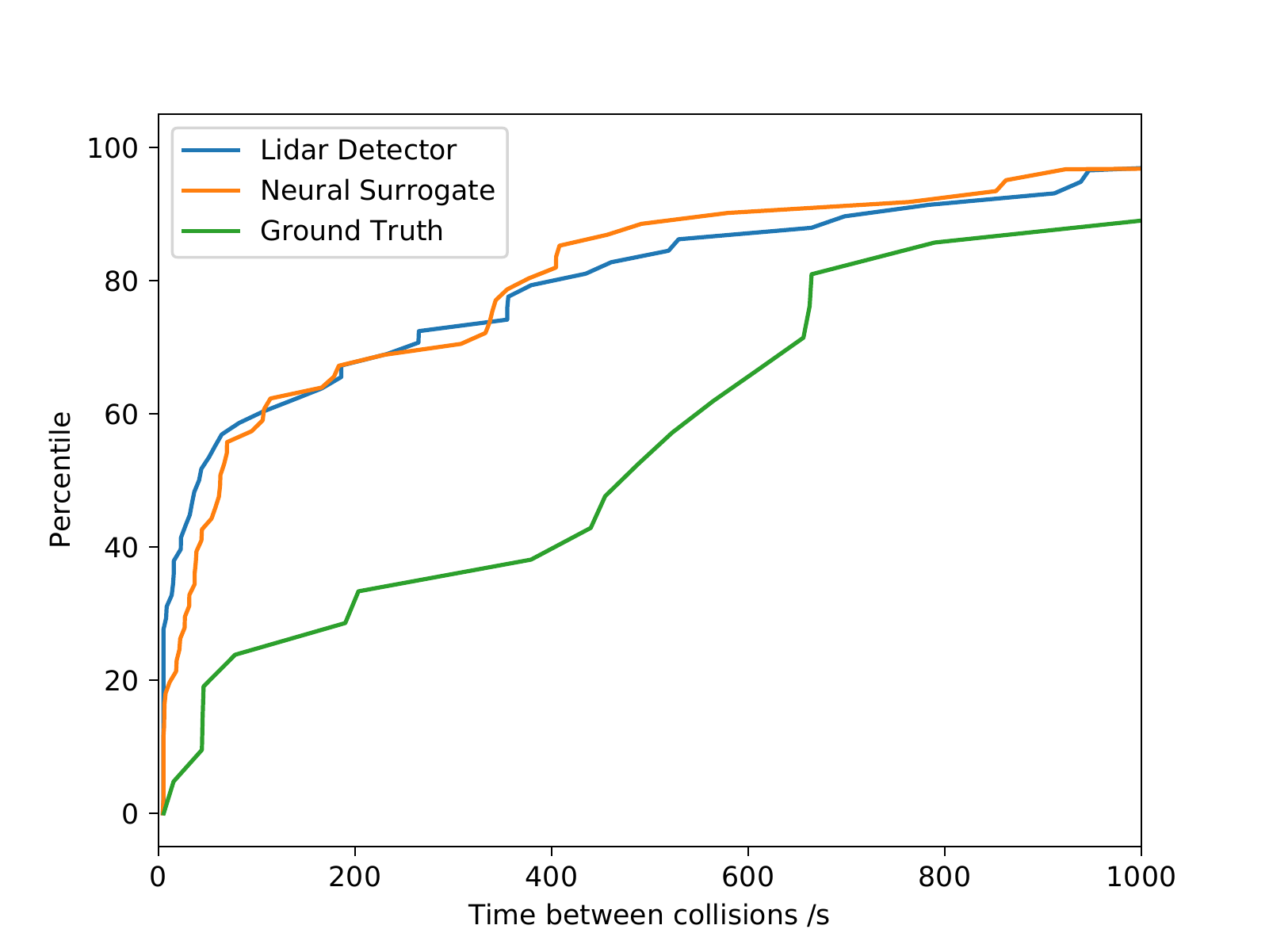}
        \caption{Cumulative distribution of the time between collisions for Carla leaderboard evaluation for NS, GT, and the CenterPoint. }
        \label{fig:time_between_collisions}
\end{minipage}
\end{figure}

\paragraph{Results:}
Metrics on the train and test set \textit{relative to the ground-truth} are shown in Figure~\ref{fig:test_2} for both the NS and the detector. 
The NS closely mimics the behaviour of the CenterPoint detector with respect to these metrics.
Figure~\ref{fig:test_comparitive_2} shows similar metrics computed for the NS \textit{relative to the detector}.
Again, the NS performs well on all metrics, particularly for objects at lower distances which are usually the most influential on the behaviour of ego.

Metrics used for Carla leaderboard evaluation are summarised in Table~\ref{tab:exp_2_results}.
Since the NS does not model false positive detections, the route completion is lower in some scenarios where a false positive LiDAR detection of street furniture confuses the planner, which does not happen for the NS or the GT.
We leave the surrogate modelling of false positive detections as a future research direction.
Figure~\ref{fig:time_between_collisions} shows cumulative distribution functions of the time between collisions; NS is \textit{clearly more similar} to the LiDAR detector than the ground truth.
The NS captures the time between collisions for the CenterPoint model much more effectively than the GT, and at a fraction of the cost of running the high-fidelity simulator and the CenterPoint detector.
Appendix~\ref{sec:example_errors} shows examples of LiDAR detector errors reproduced by the surrogate model.

\begin{table}
	\caption{Carla leaderboard evaluation metrics.}
	\centering
	\scalebox{0.82}{
		\begin{tabular}{@{}l@{\hspace{4pt}}l@{\hspace{4pt}}l@{\hspace{4pt}}l@{\hspace{4pt}}l@{}}
		\toprule
			 & LiDAR Det. & NS (our) & GT \\
			 & (CenterPoint) & &  \\
			\midrule
			\# Ped. Collisions & 1 & 1 & 1 \\
			\# Veh. Collisions & 60 & 63 & 23 \\
			Avg Completion (\%) & 81.870 & 93.933 & 91.373 \\
			Median time between collisions (seconds) & 41.250 & 62.800 & 470.950 \\
			\bottomrule
		\end{tabular}
		}
	\label{tab:exp_2_results}
\end{table}

%% file: conclusions.tex
\section{Conclusions}\label{conclusions}
In this paper, we proposed an approach to enable efficient large-scale testing of complex perceptual tasks in a simulated environment.
We showed that it is possible to create an efficient surrogate model
corresponding to heavy-compute components (for example, the backbone task of detecting objects) of a complex task such that the input now is much lower-dimensional, and the inference is multiple times faster.
We provided extensive analysis to show that such surrogate models, while showing similar behaviour to their heavy-compute counterparts when compared using variety of metrics (precision, recall, trajectory similarity, etc.), were multiple times faster as well. 

In future work, more complex surrogate models could be designed which behave more similarly to the baseline model, and the effect of surrogate model uncertainty and calibration studied.
In addition modelling of false positive detections could be added to the surrogate model. 
This is a more complex problem than false negative detections as additional non-obvious context information will need to be added to the salient variables, because false positives are often caused by factors external to the agents present in the scene. 

%% file: appendix.tex
\onecolumn

\section{Network Architecture}
\label{sec:architecture}
\begin{figure}[h]
	\centering
	\includegraphics[width=0.3\linewidth]{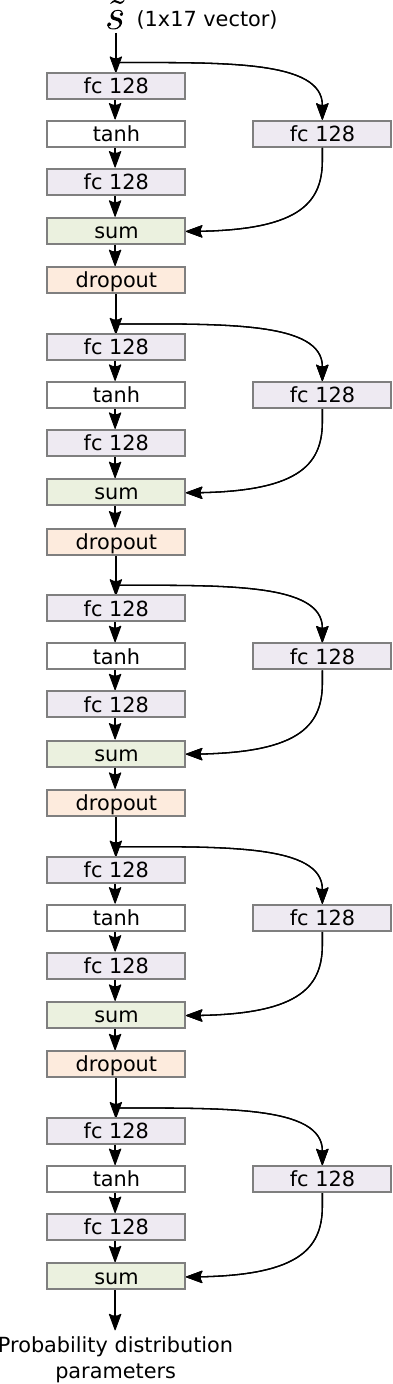}
	\caption{Diagram of architecture used for Neural Surrogate. Fully connected layers are shown as fc blocks where the number represents the output channel size.}
	\label{fig:network-architecture}
\end{figure}

Our network architecture for the surrogate is a multi-layered fully-connected network with skip connections, and dropout layers between `skip blocks' (similar to a ResNet \cite{he2016deep}), which is shown in the Figure~\ref{fig:network-architecture}.
The final layer of the network outputs the parameters of the underlying probability distributions.

\section{Experimental Hyperparameters}
\label{sec:hyperparams}

The hyperparameters used to train the surrogate models in the ACC experiment are shown in Table~\ref{tab:hyper}.
Adam optimiser was used for all training \cite{kingma2014adam}.
Hyperparameters not shown were set to default values.
The hyperparameters were selected by manual tuning.

\begin{table}[h!]

\centering

\caption{Hyperparameters used to train the surrogate models in ACC experiment.}
\begin{tabular}{l l l l }
\toprule
  & Neural Surrogate & Logistic Regression & Gaussian Fuzzer  \\ 
 \midrule
 Learning Rate & $1\mathrm{e}{-3}$ & $1\mathrm{e}{-2}$ & $1\mathrm{e}{-2}$ \\
 $N_\text{iterations}$ & 20000 & 3000 & 3000 \\
 Dropout Rate & 0.3 & N/A & N/A \\
 Batch Size & $128 \times 128$ & $16 \times 128$ & $16 \times 128$ \\
\bottomrule
\end{tabular}
\label{tab:hyper}
\end{table}

The hyperparameters used to train the neural network for the carla leaderboard evaluation are shown in Table~\ref{tab:hyper2}.

\begin{table}[h!]

\caption{Hyperparameters used to train the surrogate models in Carla leaderboard evaluation.}
\centering
\begin{tabular}{l l }
\toprule
  & Neural Surrogate  \\ 
\midrule
 Learning Rate & $1\mathrm{e}{-4}$ \\
 $N_\text{iterations}$ & 50000 \\
 Dropout Rate & 0.0 \\
 Batch Size & 12000  \\
 $\gamma$ (Focal Loss) & 2.0 \\
 $\alpha$ (Focal Loss) & 0.5 \\
 \bottomrule
\end{tabular}
\label{tab:hyper2}
\end{table}

\section{Carla Simulator Configuration}
\label{sec:sim_config}

The settings for the Carla lidar sensor are shown in Table~\ref{tab:lidar}.

\begin{table}[h!]
\caption{Lidar sensor configuration. The configuration is set to be approximately equal to an Ouster lidar sensor \cite{ouster}.}
\centering
\begin{tabular}{l l}
\toprule
 Property & Value  \\ 
 \midrule
 Sensor Position & $x=0.9, y=0.0, z=1.8$ \\  
 Sensor Orientation & roll = 0.0, pitch = 0.0, yaw = 0.0 \\
 Sensor Type & \texttt{sensor.lidar.ray\_cast} \\
 Rotation Frequency & 20 \\
 Channels & 128 \\
 Points per second & 2 620 000 \\
 Upper Field of View & 11.25 \\
 Lower Field of View & -11.25 \\
 Range & 100 \\
 Atmosphere attenuation rate & 0.004 \\
 Noise stddev & 0.1 \\
 Dropoff general rate & 0.45 \\
 Dropoff intensity limit & 0.8 \\
 Dropoff zero intensity & 0.9     \\
 \bottomrule
\end{tabular}

\label{tab:lidar}
\end{table}

\section{Planners}            

This appendix describes the ACC and \texttt{BasicAgent} planners used in the experiments in Section~\ref{sec:sim_experiment} and Section~\ref{sec:leaderboard_evaluation} respectively.

\subsection{ACC Planner Pseudo Code}
\label{sec:planner_code}

A PID controller is used in combination with the planner in Listing~\ref{code_planner} to control the vehicle throttle and brake.
The planner accelerates ego to a maximum velocity unless a slow moving vehicle is detected in the same lane as ego, in which case ego will attempt to decelerate so that ego's velocity matches that of the slow moving vehicle.
If the ego is closer than 0.1 metres to the slow moving vehicle, then it applies emergency braking.

\begin{listing*}[h]
\caption{Planner Pseudo Code}
\begin{minted}{python}
slow_threshold = 5
safe_stop_headway = 15
forward_horizon = 100
cruise_speed = 13.9
last_target_speed = cruise_speed
lane_width = 4.5
time_threshold = 0.5
timestep = 0.05

in_lane = filter(lambda x: abs(x.position.y) < lane_width / 2, objects)
within_horizon = filter(lambda x: x.position.x < forward_horizon, in_lane)
slow = list(filter(lambda x: x.speed < slow_threshold, within_horizon))

if len(slow) > 0:
    closest_agent = sorted(slow, key=lambda x: x.position.x)[0]

    distance_to_closest = distance_to_closest_agent_vertex(closest_agent)
    distance_to_safe = distance_to_closest - safe_stop_headway

    if distance_to_safe < 0.1:
        target_speed = 0
    else:
        speed_of_closest = closest_agent.speed

        if speed_of_closest < 0.5:
            speed_of_closest = 0

        accel = (speed_of_closest ** 2 - last_target_speed ** 2) / (2 * distance_to_safe)
        target_speed = last_target_speed + accel * timestep

else:
    target_speed = last_target_speed + target_accel * timestep
\end{minted}
\label{code_planner}
\end{listing*}

\subsection{BasicAgent Modifications}
\label{sec:basicagent}
The \texttt{BasicAgent} planner uses a PID controller to accelerate the vehicle to a maximum speed, and stops only if a vehicle within a semicircle of specific radius in front of ego is detected where the vehicle's centre is in the same lane as ego. 
We made changes to the \texttt{BasicAgent} planner to improve its performance.
We modified \texttt{BasicAgent} to avoid pedestrian collisions and brake when the corner of a vehicle is inside a rectangle of lane width in front of the ego such that the vehicle's lane is the same as one of ego's future lanes.
Also, the \texttt{BasicAgent} was modified to drive slower close to junctions.

\section{Surrogate model error distributions}
\label{sec:error_distributions}

\begin{figure}
\centering
\begin{subfigure}[b]{0.45\linewidth}
         \centering
         \includegraphics[width=\linewidth, trim=2cm 2.5cm 1cm 2.7cm,clip]{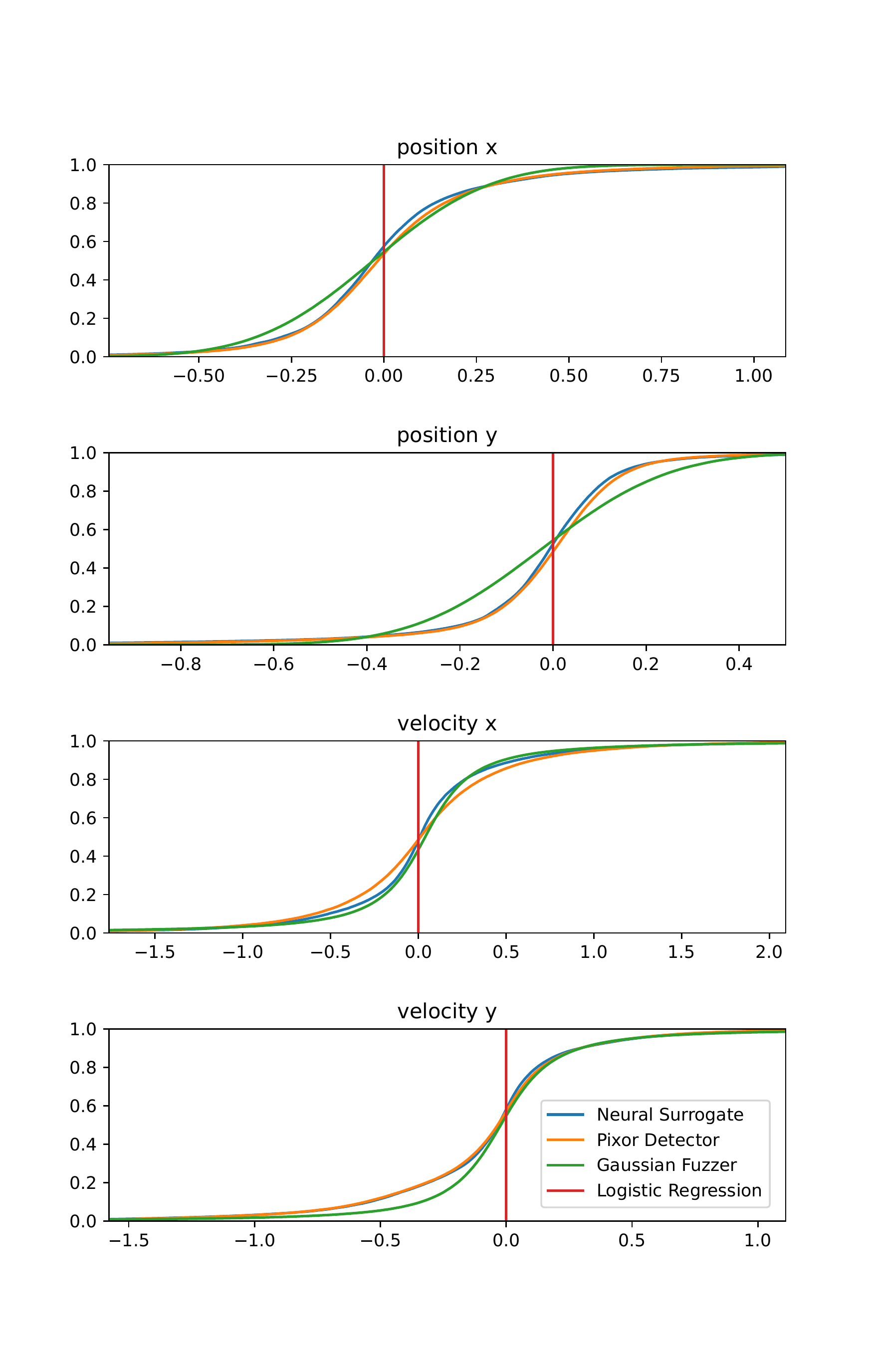}
         \caption{Train set}
\end{subfigure}
\begin{subfigure}[b]{0.45\linewidth}
     \centering
         \includegraphics[width=\linewidth, trim=2cm 2.5cm 1cm 2.7cm,clip]{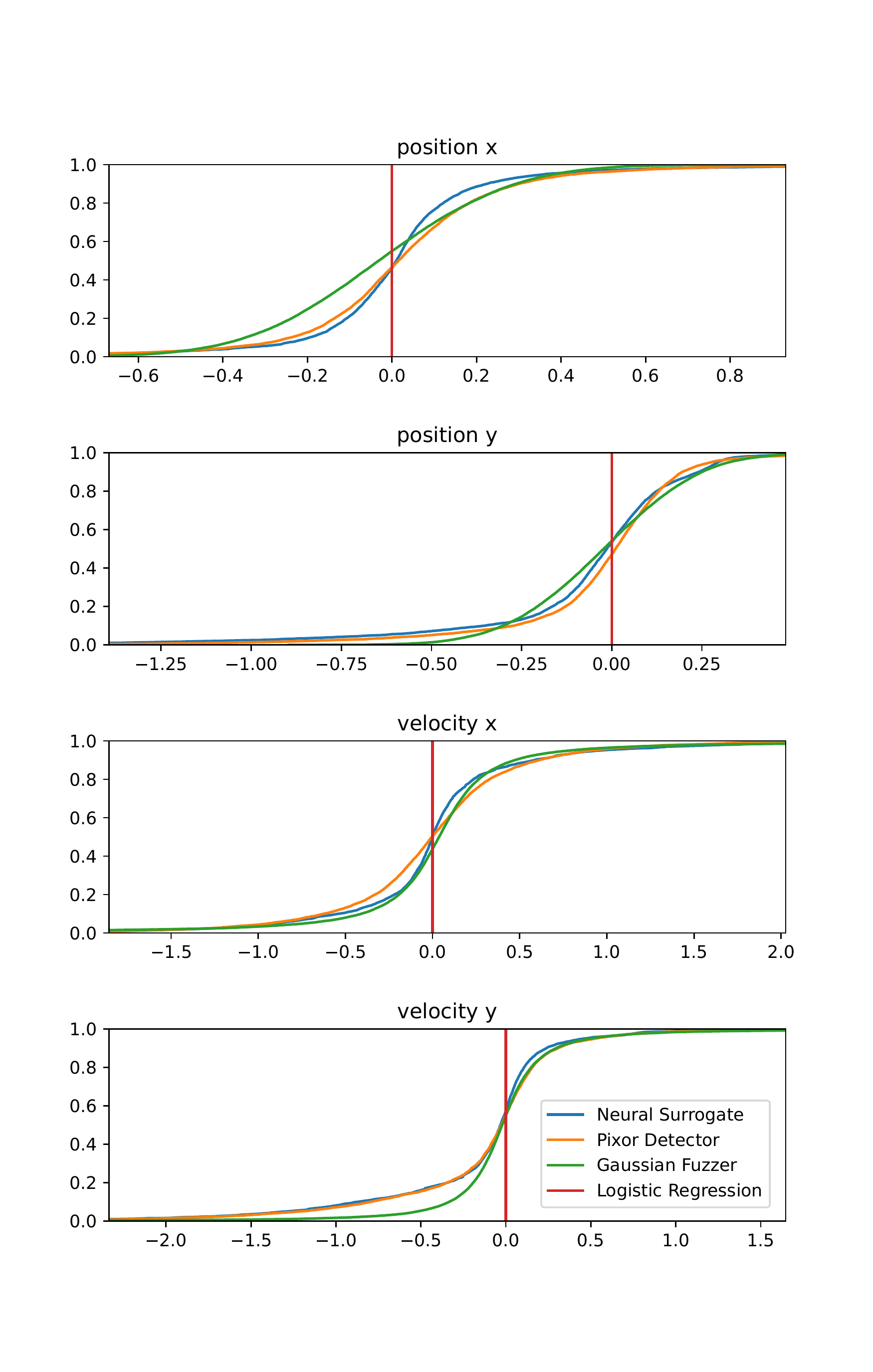}
         \caption{Test set}
\end{subfigure}
        \caption{Surrogate model fitted error distributions on train and test set. Note that in all cases the Neural Surrogate (our) distribution is closer than any of the other baselines to the distribution of the PIXOR Detector being approximated. Although the fit of the models to the training set appears similar from these plots, the performance of a planner downstream from these models is both visually and quantifiably different.}
        \label{fig:error_plots}
\end{figure}

Empirical cumulative distribution functions for the positional and velocity error predicted by each surrogate model and the true detector relative to the true object locations are shown in Figure~\ref{fig:error_plots}.

\section{Ego Behaviour Diagnostics}
\label{sec:diagnostic_plots}

\begin{figure}
	\centering
	\includegraphics[width=0.7\linewidth]{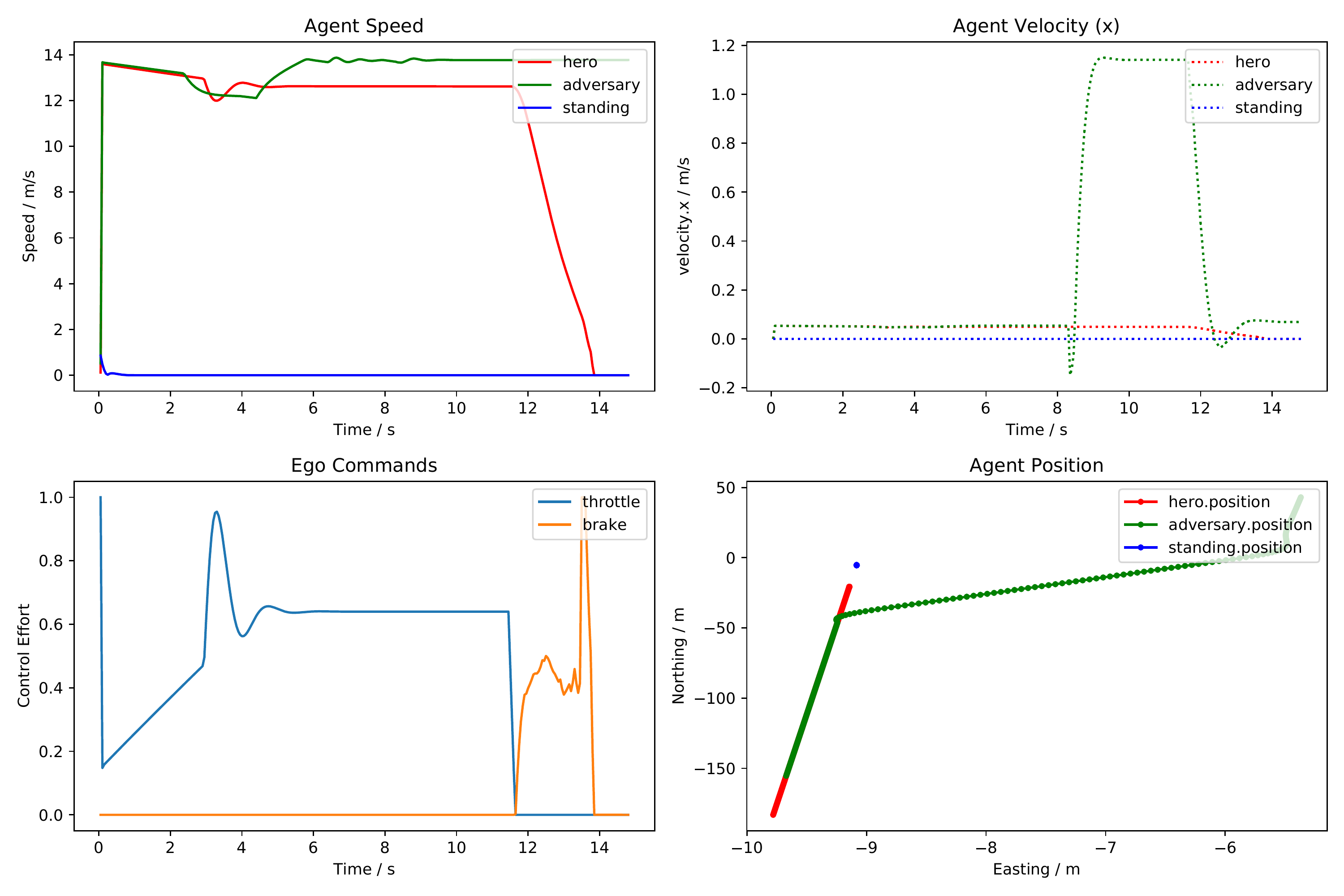}
	\caption{Diagnostics for simulation with the full upstream detector task.}
	\label{fig:diagnostic_1}
\end{figure}

\begin{figure}
	\centering

	\includegraphics[width=0.7\linewidth]{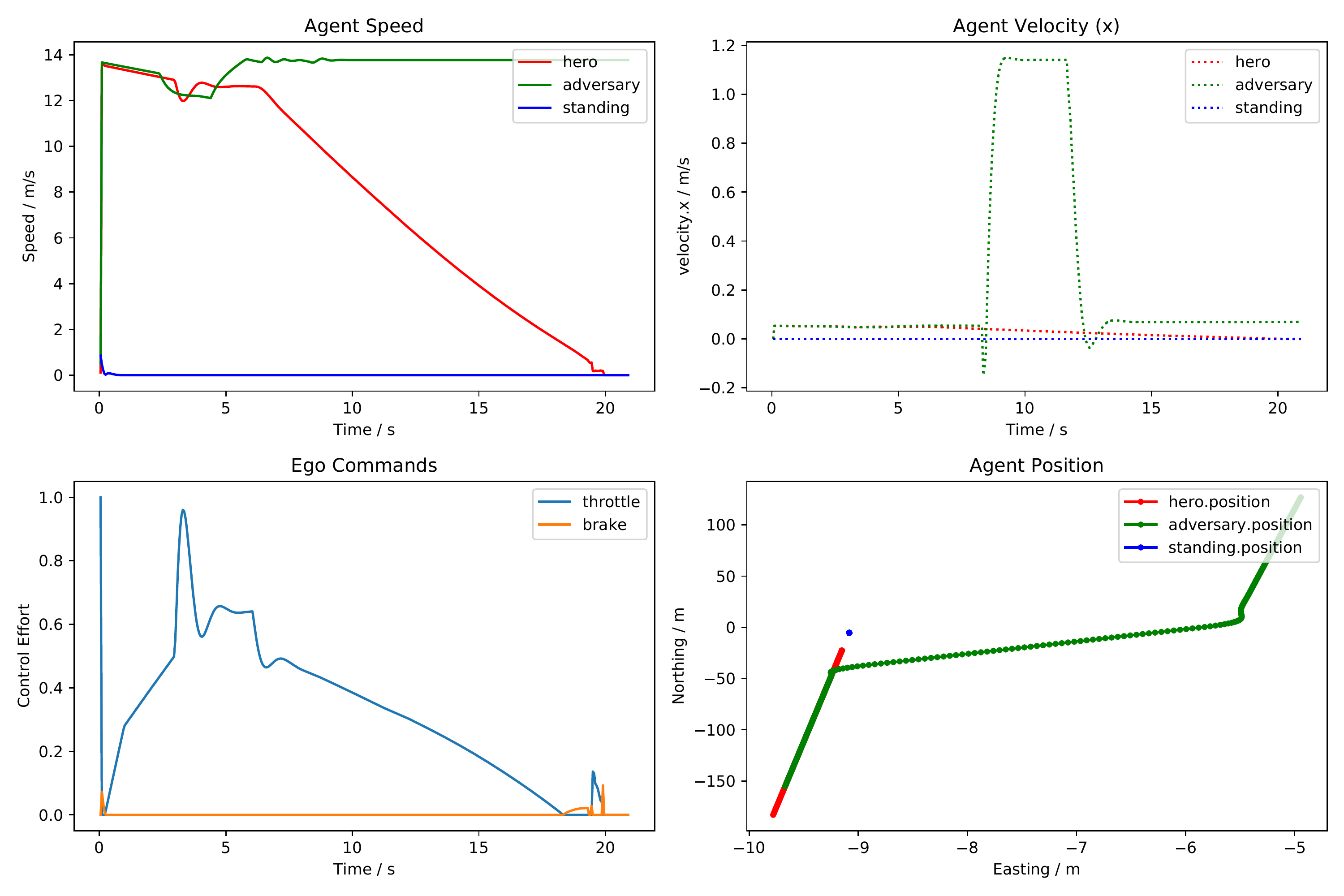}
	\caption{Diagnostics for simulation with the upstream detector outputs substituted for ground truth values.}
	\label{fig:diagnostic_2}
\end{figure}

\begin{figure}
	\centering
	
	\includegraphics[width=0.7\linewidth]{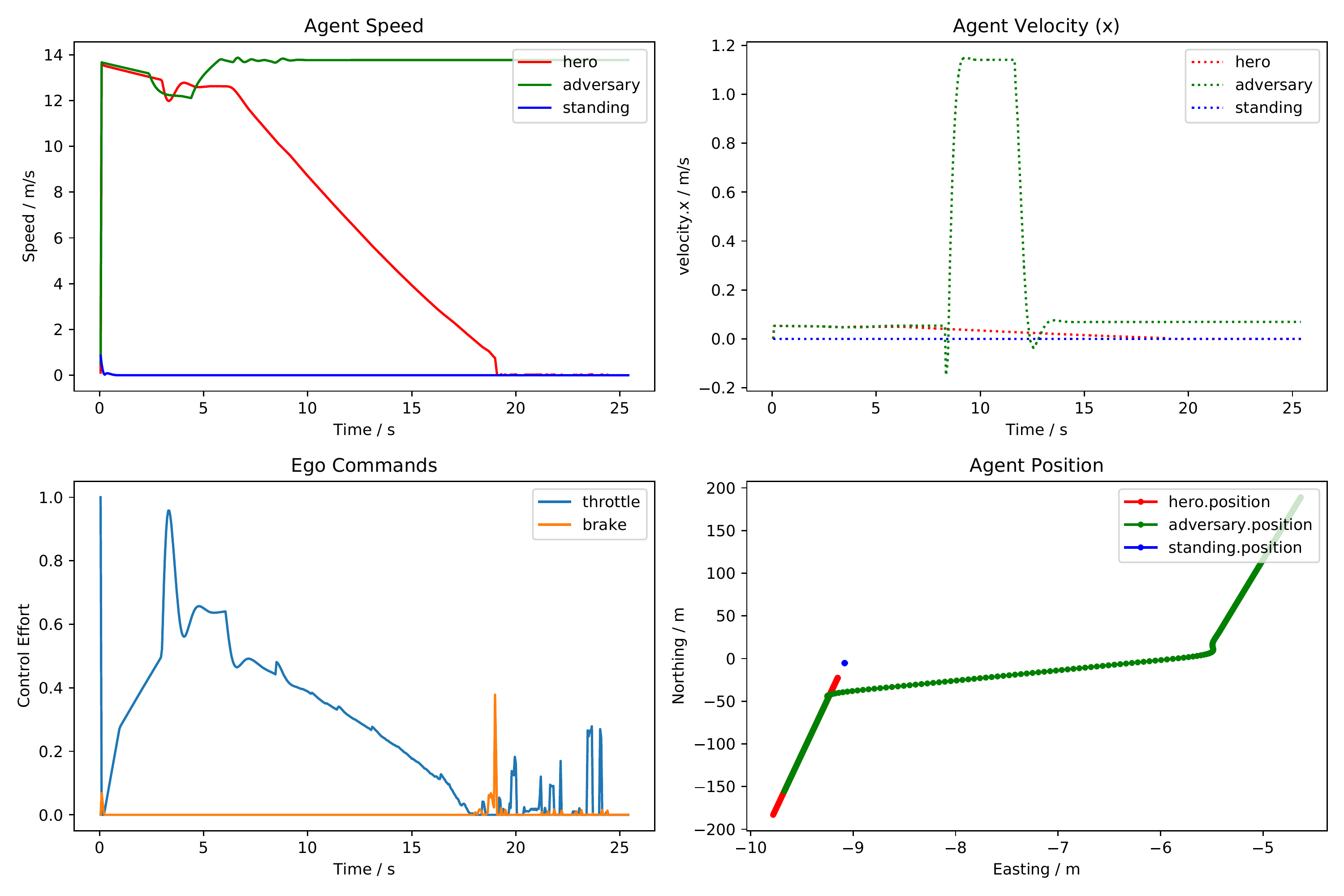}
	\caption{Diagnostics for simulation with the upstream detector outputs substituted for values generated by a simple `Fuzzer' surrogate model.}
	\label{fig:diagnostic_3}
\end{figure}

\begin{figure}
	\centering
	
	\includegraphics[width=0.7\linewidth]{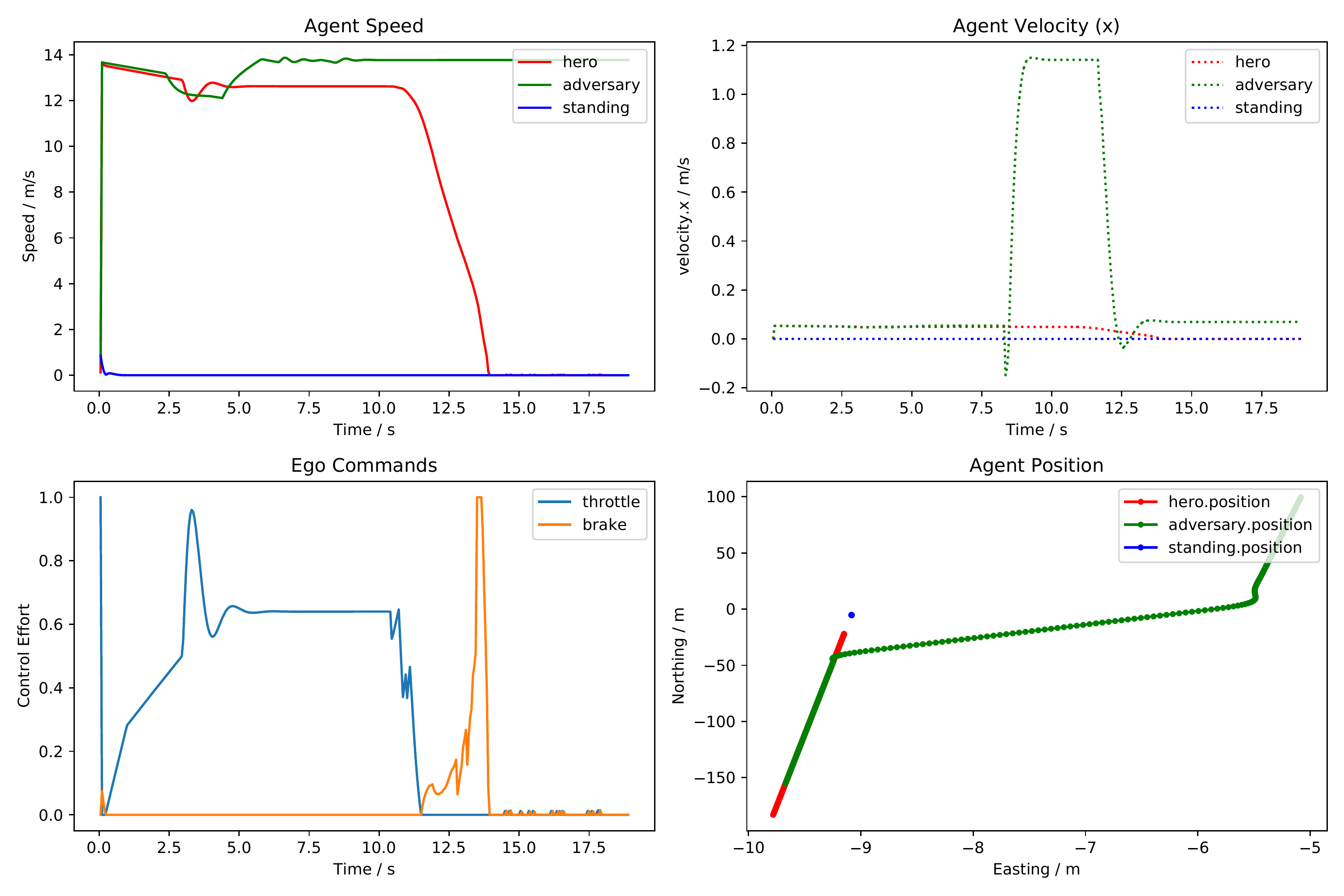}
		\caption{Diagnostics for simulation with the upstream detector outputs substituted for values generated by a neural network surrogate model.}
		\label{fig:diagnostic_4}
\end{figure}

\begin{figure}
	\centering
	
	\includegraphics[width=0.7\linewidth]{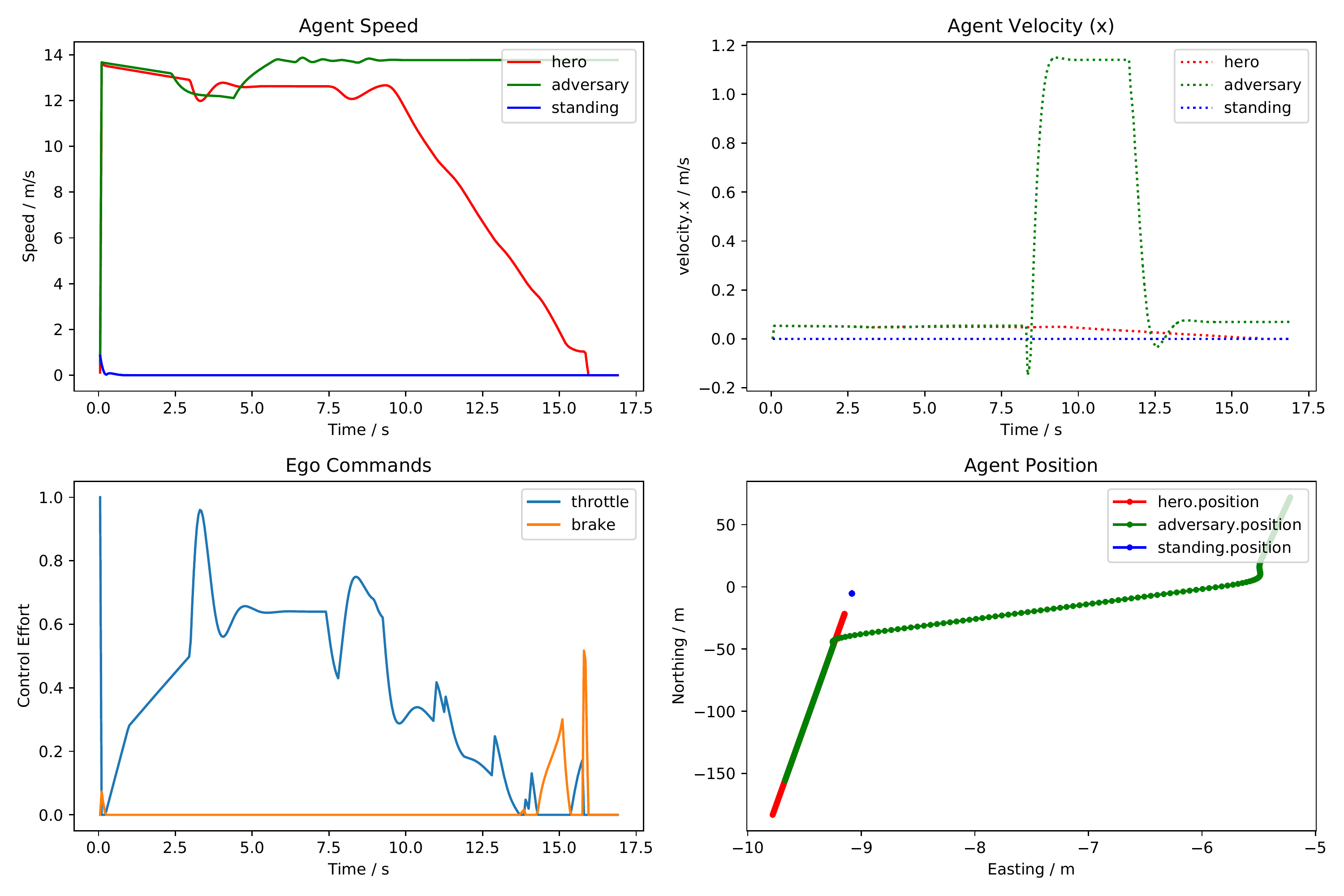}
		\caption{Diagnostics for simulation with the upstream detector outputs substituted for values generated by a logistic regression surrogate model.}
		\label{fig:diagnostic_5}
\end{figure}

The diagnostics for the ACC experiment are shown in Figures~\ref{fig:diagnostic_1}, \ref{fig:diagnostic_2}, \ref{fig:diagnostic_3}, \ref{fig:diagnostic_4} and \ref{fig:diagnostic_5}.

\section{Comparison with Planner KL Divergence}
\label{sec:pkl-comparison}

In this appendix, we explain the relationship between the mean Euclidean norm metric, and the Planner KL divergence metric proposed by \citet{philion2020learning}.

The KL divergence between the plan produced by an agent planning based on a detector and a surrogate model is given by:
\begin{equation}
	\text{P-KL} = \expectation_{p(\bfz^1 | f(\bfx))} \left( \log{\frac{\prod_{t=1}^{t=T} p(\bfz_t^1 | f(\bfx))}{\prod_{t=1}^{t=T} \int  p(\bfz_t^1 | \bfy) p(\bfy | \tilde{\bfs}) d\bfy}} \right)   , \label{eqn:pkl-def}
\end{equation}
where $\bfz_t^1$ is the position of ego at timestamp $t$, $\bfz^1 = \{\bfz_1^1,...,\bfz_T^1\}$, $f: \bfx \mapsto \bfy$ is the detector, $p(\bfz_t^1 | \bfy)$ is the probabilistic planner, and $p(\bfy | \tilde{\bfs})$ is the probability distribution associated with the surrogate model for the detector, i.e. $\tilde{f}(\bfy | \tilde{\bfs}) \sim p(\bfy | \tilde{\bfs})$, which produces detections $\bfy \in \calY $ from salient variables $\tilde{\bfs}$.

The planner in our case is deterministic, so 
\begin{equation}
p(\bfz_t^1 | \bfy) =  \indicator_{\{\bfz_t^1 = g(\bfy)\}}
\end{equation}
with the deterministic planning function $g(\bfy)$, which can be used to rewrite Equation~\ref{eqn:pkl-def} as 
\begin{equation}
	\text{P-KL} = - \sum_{t=1}^{t=T} \left. \log \left(  \int  p(\bfz_t^1 | \bfy) p(\bfy | \tilde{\bfs}) d\bfy  \right) \right\vert_{\bfz^1 = g(f(\bfx))}.
\end{equation}
Approximating the integral $ \int  p(\bfz_t^1 | \bfy) p(\bfy | \tilde{\bfs}) d\bfy$ with a kernel density estimator with bandwidth $h$ obtained by sampling $n$ times from $p(\bfy | \tilde{\bfs})$ yields
\begin{equation}
	\text{P-KL} = - \sum_{t=1}^{t=T} \left. \log \left( \frac{1}{nh} \sum_\bfy K\left( \frac{\bfz_t^1 - g(\bfy)}{h} \right) \right) \right\vert_{\bfz^1 = g(f(\bfx))},
\end{equation}
after which Jensen's inequality can be applied to obtain
\begin{equation}
	\text{P-KL} \le -\frac{1}{n} \sum_{t=1}^{t=T} \left.   \sum_\bfy \log \left( \frac{1}{h} K\left( \frac{\bfz_t^1 - g(\bfy)}{h} \right) \right) \right\vert_{\bfz^1 = g(f(\bfx))},
\end{equation}
which with a Gaussian kernel is equivalent to 
\begin{equation}
\text{P-KL} \le 
\left. \frac{1}{n} \sum_{t=1}^{t=T}  \sum_\bfy  \left(\frac{(\bfz_t^1 - g(\bfy))^2}{2h^2} + \log \left( \sqrt{2 \pi} h \right) \right) \right\vert_{\bfz^1 = g(f(\bfx))},
\end{equation}
which is equal to the mean Euclidean norm metric (Eqn.~\ref{eqn:euclidean}) up to a constant and scaling factor, when the surrogate model is sampled repeatedly and the average taken.

\section{Example detection errors from simulation}
\label{sec:example_errors}
\begin{figure}
     \centering
          \begin{subfigure}[b]{0.3\linewidth}
         \centering
         \includegraphics[width=\linewidth]{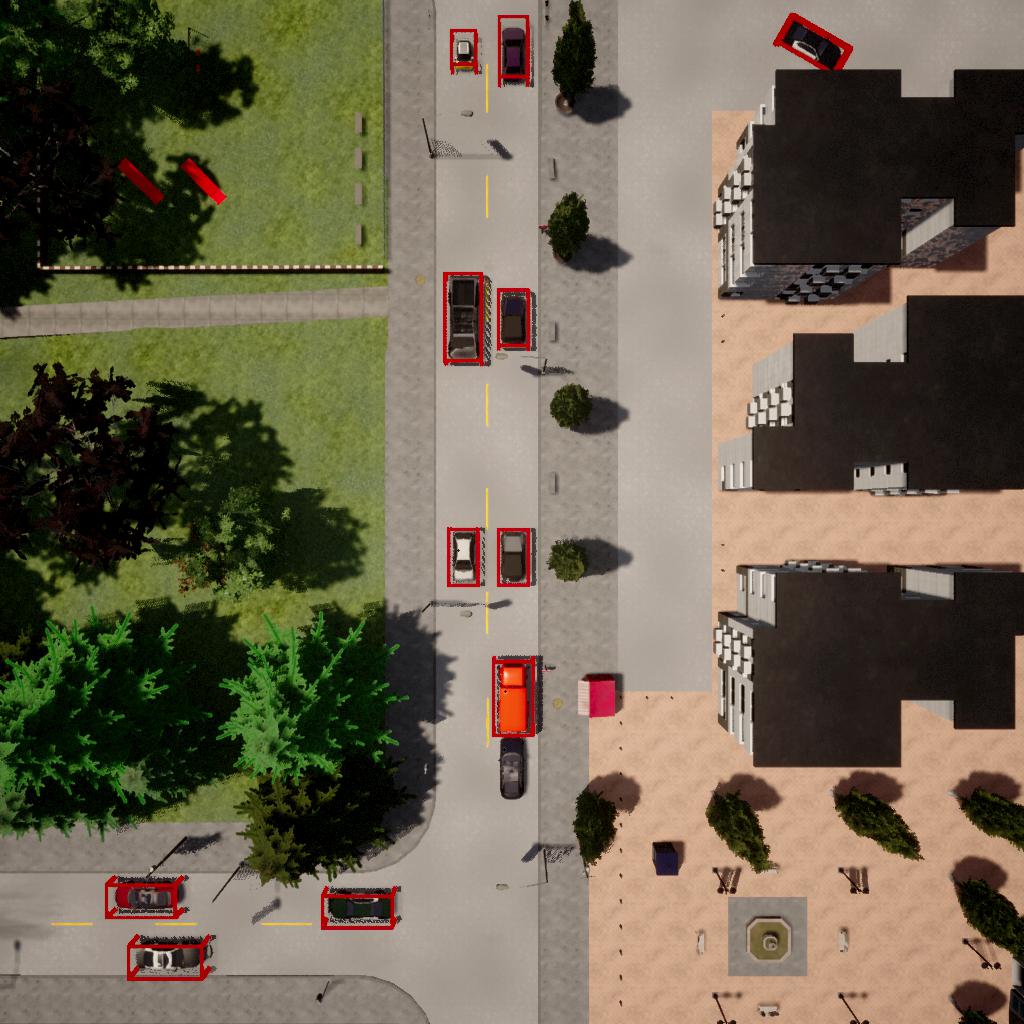}
         \caption{Ground Truth}
     \end{subfigure}
     \begin{subfigure}[b]{0.3\linewidth}
         \centering
         \includegraphics[width=\linewidth]{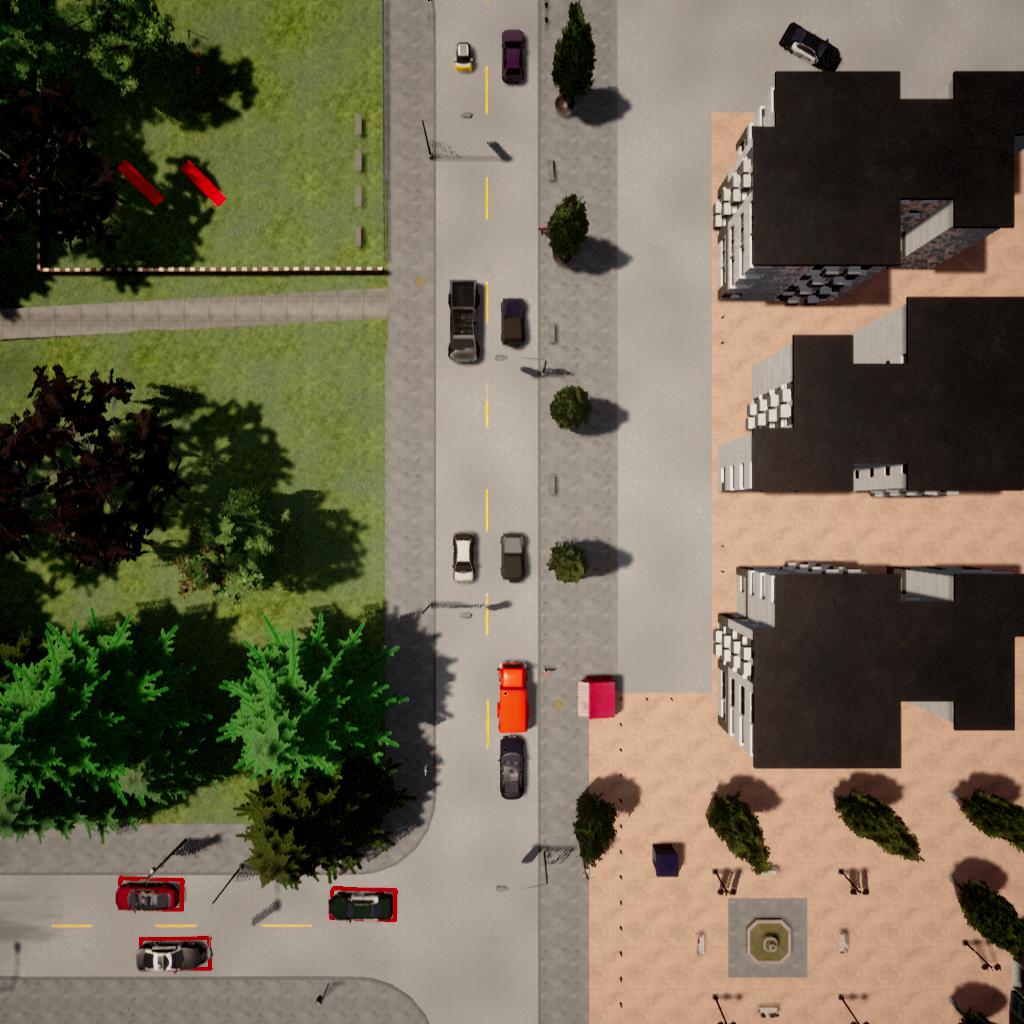}
         \caption{Neural Surrogate}
     \end{subfigure}
     \begin{subfigure}[b]{0.3\linewidth}
         \centering
         \includegraphics[width=\linewidth]{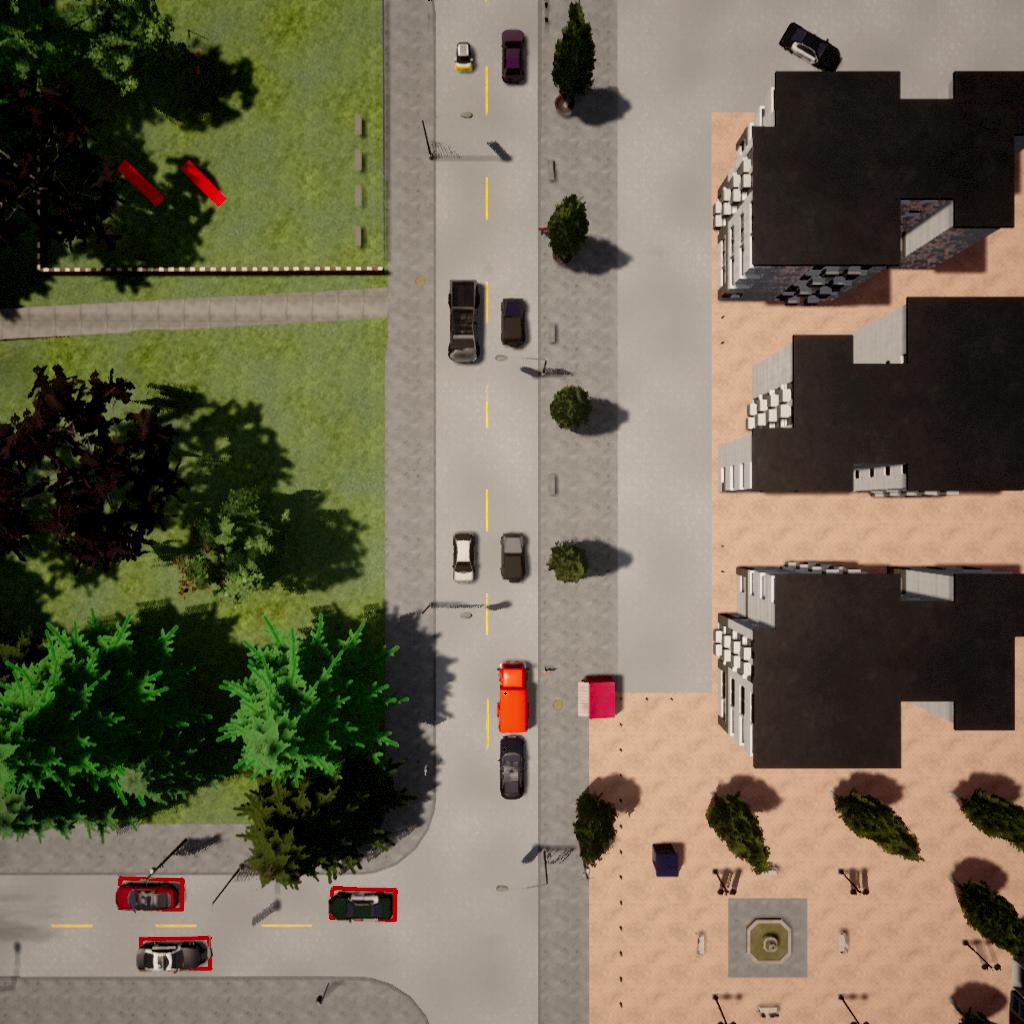}
         \caption{Lidar Detector}
     \end{subfigure}
        \caption{Example of vehicle false negative detection causing collision in Route 2. Detected objects are shown in red bounding boxes. Ego is the black vehicle colliding with the rear of the red van.}
        \label{fig:example_scenes_1}
\end{figure}

\begin{figure}
     \centering
          \begin{subfigure}[b]{0.3\linewidth}
         \centering
         \includegraphics[width=\linewidth]{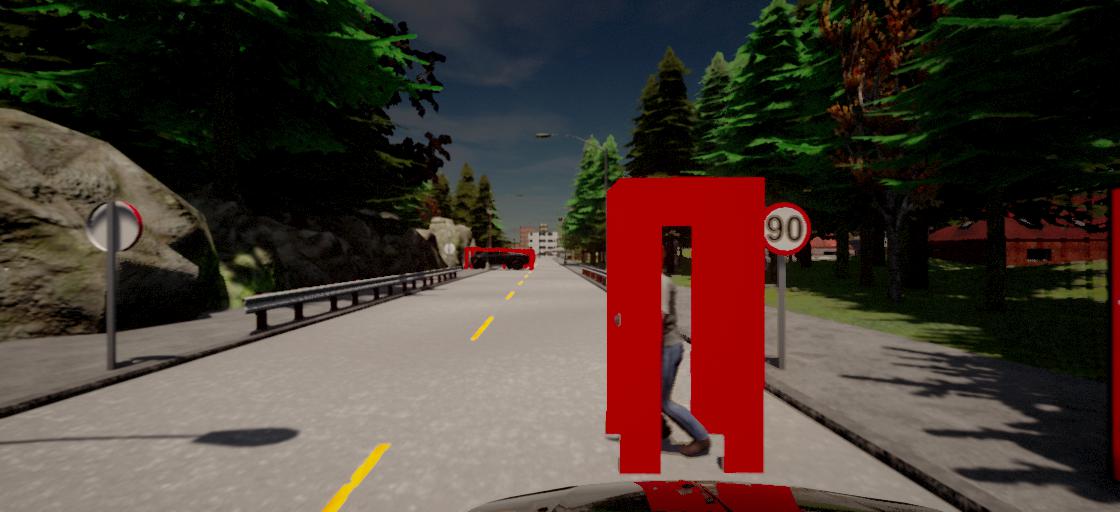}
         \caption{Ground Truth}
     \end{subfigure}
     \begin{subfigure}[b]{0.3\linewidth}
         \centering
         \includegraphics[width=\linewidth]{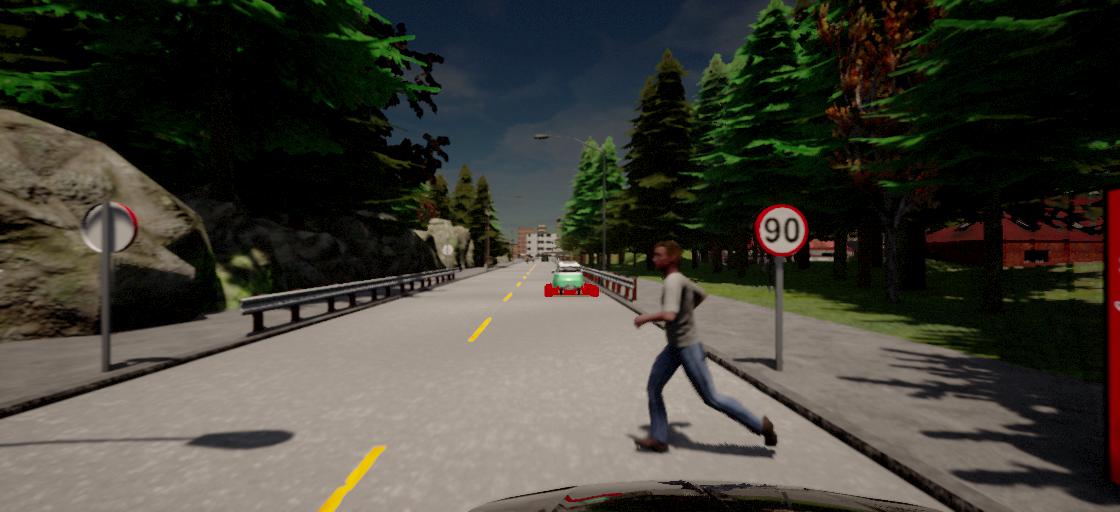}
         \caption{Neural Surrogate}
     \end{subfigure}
     \begin{subfigure}[b]{0.3\linewidth}
         \centering
         \includegraphics[width=\linewidth]{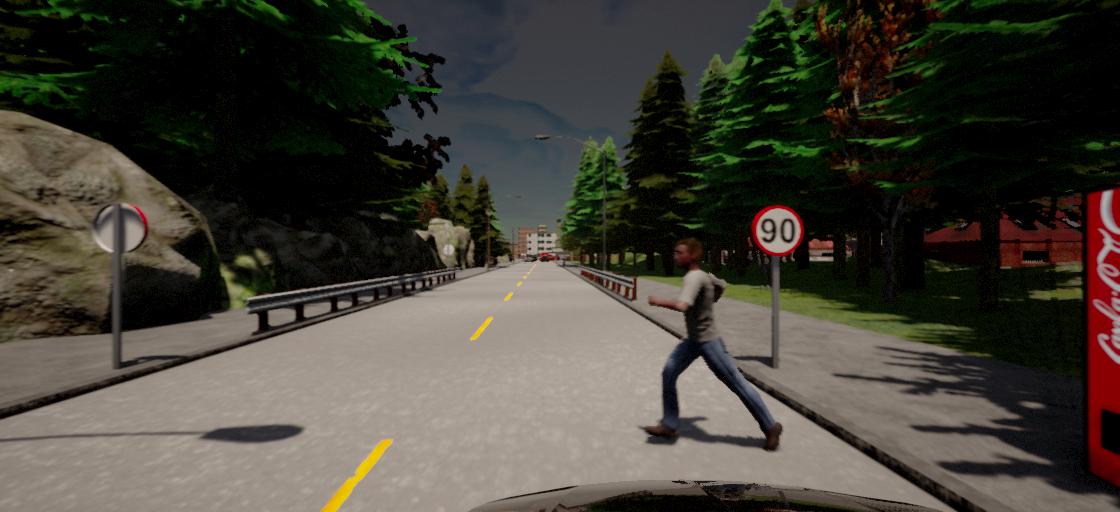}
         \caption{Lidar Detector}
     \end{subfigure}
        \caption{Example of pedestrian false negative detection causing collision in Route 1. Detected objects are shown in red bounding boxes.}
        \label{fig:example_scenes_2}
\end{figure}

\begin{figure}
     \centering
          \begin{subfigure}[b]{0.3\linewidth}
         \centering
         \includegraphics[width=\linewidth]{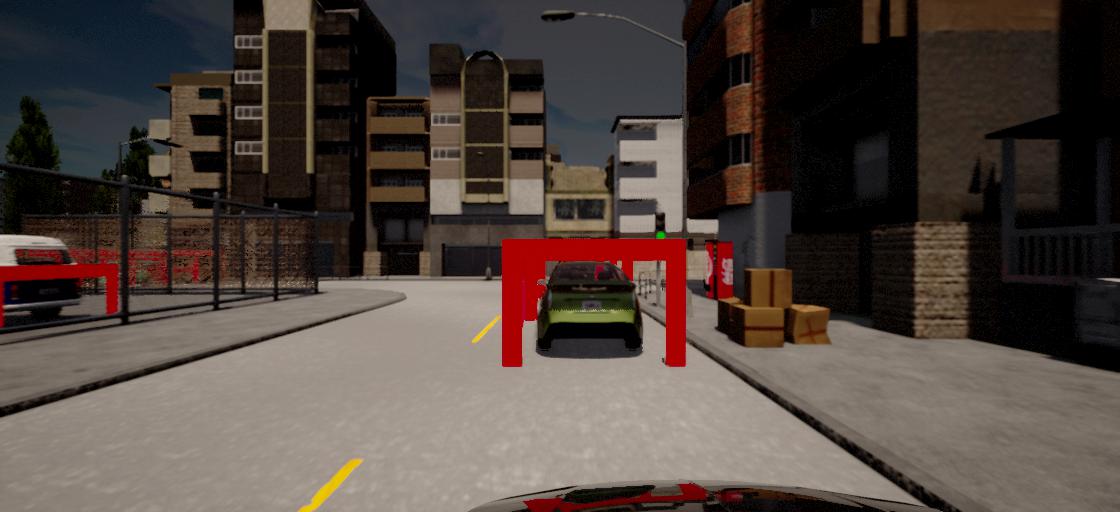}
         \caption{Ground Truth}
     \end{subfigure}
     \begin{subfigure}[b]{0.3\linewidth}
         \centering
         \includegraphics[width=\linewidth]{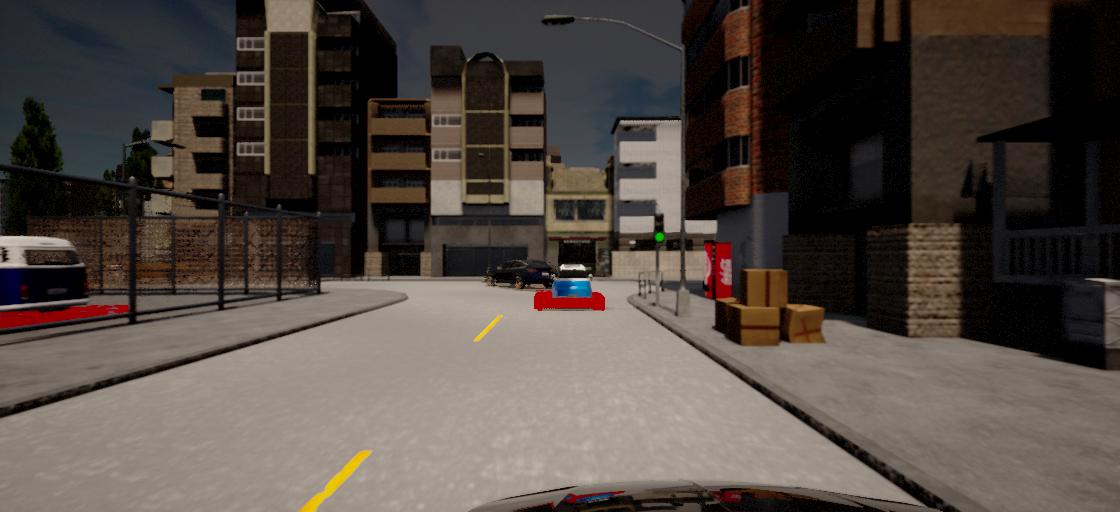}
         \caption{Neural Surrogate}
     \end{subfigure}
     \begin{subfigure}[b]{0.3\linewidth}
         \centering
         \includegraphics[width=\linewidth]{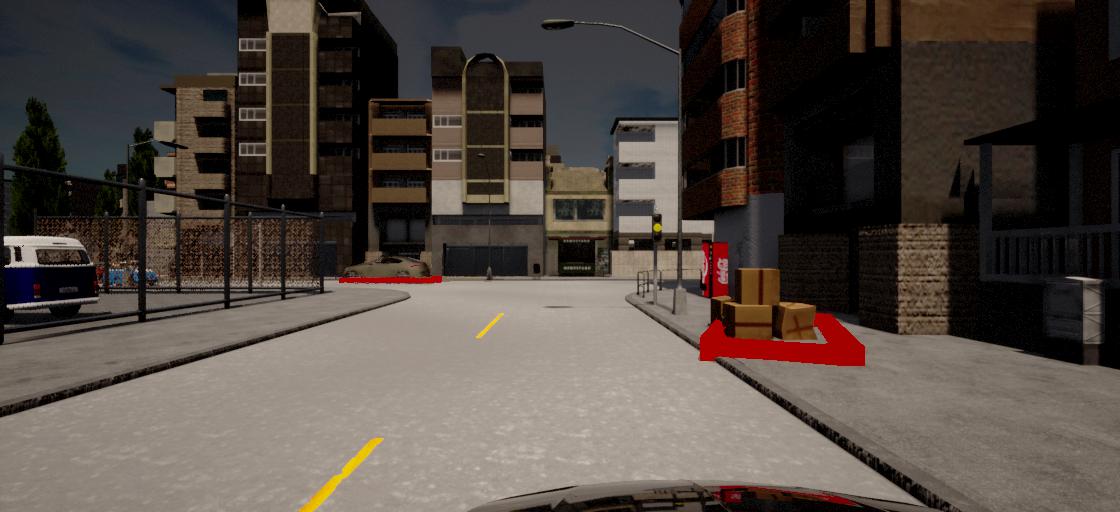}
         \caption{Lidar Detector}
     \end{subfigure}
        \caption{Example of false positive detection with a corner of the detected object in Ego's lane causing ego to stop in Route 0. Detected objects are shown in red bounding boxes.}
        \label{fig:example_scenes_3}
\end{figure}

Figures~\ref{fig:example_scenes_1}, \ref{fig:example_scenes_2} and \ref{fig:example_scenes_3} show examples of lidar detector errors which cause collisions which are reproduced by the surrogate model, and would not be reproduced if ground truth values were used in place of the backbone model in simulation.

